%% file: DeepFeatureConsistency.tex
\documentclass[a4paper,fleqn]{cas-dc}

\usepackage[numbers]{natbib}

\usepackage{stfloats}
\usepackage{caption}
\usepackage{subcaption}
\usepackage{graphicx}
\usepackage{float}
\usepackage{gensymb}
\usepackage{booktabs}
\usepackage{multirow}
\usepackage{geometry}
\usepackage{blindtext}
\usepackage{makecell}
\usepackage{stmaryrd}
\usepackage{textcomp}
\usepackage{wrapfig}
\usepackage{xparse}
\usepackage{morewrites}
\usepackage{xkeyval}%
\usepackage{epstopdf}
\def\tsc#1{\csdef{#1}{\textsc{\lowercase{#1}}\xspace}}
\tsc{WGM}
\tsc{QE}

\begin{document}
\let\WriteBookmarks\relax
\def\floatpagepagefraction{1}
\def\textpagefraction{.001}

\shorttitle{Deep Feature Consistency for Robust  Point Cloud Robust Registration}    

\shortauthors{Z. Xu et al.}  

\title [mode = title]{DFC: Deep Feature Consistency for Robust  Point Cloud Registration}  


%

\author{Zhu~Xu}[type=editor,
    style=chinese,
    auid=000,
    bioid=1]



\ead{xuzhu@mail.ynu.edu.cn}



\address{School of Information Science and Engineering, Yunnan University, Kunming City 650500, China}

\author{Zhengyao~Bai}[type=editor,
                    style=chinese,
                    auid=000,
                    bioid=1]

\ead{Baizhy@ynu.edu.cn}

\cortext[1]{Corresponding author}
\cormark[1]



\author{Huijie~Liu}[type=editor,
                style=chinese,
                auid=000,
                bioid=1]

\author{Qianjie~Lu}[type=editor,
                style=chinese,
                auid=000,
                bioid=1]

\author{Shenglan~Fan}[type=editor,
                style=chinese,
                auid=000,
                bioid=1]

\begin{abstract}
How to extract significant point cloud features  and estimate the pose between them remains a challenging question, due to the inherent lack of structure and ambiguous order permutation of point clouds. Despite  significant improvements in applying deep learning-based methods for most 3D computer vision tasks, such as object classification, object segmentation and point cloud registration,  the consistency between features is still not attractive in existing learning-based pipelines. In this paper, we present a novel learning-based alignment network for complex alignment scenes, titled deep feature consistency and consisting of three main modules: a multiscale graph feature merging network for converting the geometric correspondence set into high-dimensional features, a correspondence weighting module for constructing multiple candidate inlier subsets, and a Procrustes approach named deep feature matching for giving a closed-form solution to estimate the relative pose. As the most important step of the deep feature matching module, the feature consistency matrix for each inlier subset is constructed to obtain its principal vectors as the inlier likelihoods of the corresponding subset. We comprehensively validate the robustness and effectiveness of our approach on both the 3DMatch dataset and  the KITTI odometry dataset. For large indoor scenes, registration results on the 3DMatch dataset demonstrate that our method outperforms both the state-of-the-art traditional and learning-based methods. For KITTI outdoor scenes, our approach remains quite capable of lowering the transformation errors. We also explore its strong generalization capability over cross-datasets.
\end{abstract}


\begin{highlights}
\item Investigate the pairwise registration problem between two scans with local sparsity and partial correspondences.
\item Assign weights for correspondences to complete robust and efficient registration.
\item Focus on the consistency between features of the correspondences.
\item Achieve state-of-the-art performance on real-world  scenarios, compared with other advanced registration approaches. 
\end{highlights}


\maketitle
\captionsetup[figure]{name={Fig.},labelsep=period,singlelinecheck=false} 
\input{includefiles/section01.tex}
\input{includefiles/section02.tex}
\input{includefiles/section03.tex}
\input{includefiles/section04.tex}
\input{includefiles/section05.tex}
\input{includefiles/section06.tex}

\section*{Acknowledgements}
This work was supported by the Yunnan provincial major science and technology special plan projects: digitization research and application demonstration of Yunnan characteristic industry, under Grant: 202002AD080001, and the Practice\&Innovation Foundation for Professional  Degree Graduates of Yunnan University, under Grant: 2021Y168.

\bibliographystyle{model2-names}

\bibliography{cas-refs.bib}


\end{document}

%% file: includefiles/section01.tex
\section{Introduction}\label{sec01Introduction}
Point cloud registration is playing an increasingly important role in many applications, including simultaneous localization and mapping~(SLAM), 3D reconstruction and autonomous driving. Point clouds have many specific characteristics that may increase the complexity of registration problems, including local sparsity, noise caused by acquisition equipment and a large number of points. On the one hand, point cloud  sparsity and noise make it unrealistic to find correct correspondences between the source and target point clouds. On the other hand, the considerable number of points inevitably requires efficient algorithms and large computing resources.
\par 
Traditional point cloud registration pipelines~(\citeauthor{ICP1992}, \citeyear{ICP1992}; \citeauthor{GoICP2016}, \citeyear{GoICP2016}) start with a coarse initial pose obtained by odometers and iterate until the optimal condition is satisfied. However, the registration result is highly dependent on a good initial estimation, which directly tends to cause these pipelines to become stuck in local minima. To increase registration accuracy and efficiency, researchers have proposed learning-based algorithms to replace the individual parts in the classical registration pipeline, including feature descriptors~(\citeauthor{FullyConvolutionalGeometricFeatures2019}, \citeyear{FullyConvolutionalGeometricFeatures2019}; \citeauthor{DeepClosetPoint2019}, \citeyear{DeepClosetPoint2019}) and pose optimization algorithms~(\citeauthor{FastGlobalRegistration2016}, \citeyear{FastGlobalRegistration2016}; \citeauthor{GoICP2016}, \citeyear{GoICP2016}). Specifically, end-to-end registration networks, such as DCP~(\citeauthor{DeepClosetPoint2019}, \citeyear{DeepClosetPoint2019}), PointNetLK~(\citeauthor{PointNetLK2019}, \citeyear{PointNetLK2019}), and VCR-Net~(\citeauthor{VitualCorresponmdences2020}, \citeyear{VitualCorresponmdences2020}), have gradually emerged in recent years. Compared with other classical registration pipelines~(\citeauthor{ICP1992}, \citeyear{ICP1992}; \citeauthor{GoICP2016}, \citeyear{GoICP2016}; \citeauthor{RANSAC1981}, \citeyear{RANSAC1981}), the high efficiency of end-to-end neural networks has been fully verified. However, the robustness and application ability of end-to-end registration pipelines cannot achieve the expected effect, especially in some complex scenes. In summary, 3D point cloud registration is still  an extremely challenging topic in 3D computer vision due to the point cloud characteristics mentioned above. 
\par 
In this work, we present a novel method titled deep feature consistency network to jointly solve the inlier correspondences and estimate the rigid transformation in the absence of initial transformation by leveraging the feature consistency during the feature matching stage.  
\par 
The  main ideas of this paper are as follows. First, a multiscale graph feature merging network is proposed to extract the features of the correspondence set. Second, to effectively filter the outliers, we present a correspondence weighting module to estimate the confidence of each correspondence based on their features and then select a series of correspondences with a high confidence level as candidate inliers based on the confidence. These candidate inliers form a series of inlier subsets in the feature space as the input to the subsequent deep feature matching module. Finally, we present a fresh deep feature matching module, which constructs the feature consistency matrix of each inlier subset in parallel, calculates the principal vectors of the feature consistency matrix with principal component analysis~(PCA)~(\citeauthor{PCA1987}, \citeyear{PCA1987}), and then obtains the corresponding rigid transformation of each inlier subset by the weighted singular value decomposition~(SVD) optimization method. The determination of the optimal rigid transformation among the above obtained rigid transformations ultimately depends on maximizing the geometric consistency.
\par 
To summarize, the key contributions of our work are as follows. 
\par 
First, we investigate the point cloud pairwise registration problem between two fragments with local sparsity and only  a part of the correspondences.  
\par
Second, we present a novel feature embedding method called multiscale graph feature merging network. This method is used to extract the  final features of the correspondence set  and can well make full use of the geometric connection information between the nearest correspondences.
\par 
Third, we propose a deep feature matching module that is designed to predict the rigid transformation used to align the point clouds to boost the registration performance.
\par 
Experimentally, comparison results on the 3DMatch dataset reveal that our approach achieves state-of-the-art performance, compared with both classical~(\citeauthor{RANSAC1981}, \citeyear{RANSAC1981}; \citeauthor{ICP1992}, \citeyear{ICP1992}) and learning-based approaches~(\citeauthor{PointDSC2021}, \citeyear{PointDSC2021}; \citeauthor{DeepGlobalRegistration2020}, \citeyear{DeepGlobalRegistration2020}). In addition, our method shows strong generalization ability over different datasets.

%% file: includefiles/section02.tex
\section{Related Work}\label{Sec02RelatedWork}
\subsection{Feature Extraction}
Generally, we have to take measures to extract pointwise features before performing registration. There are four mainstream methods for extracting point features. The first is to convert a point cloud into a volumetric representation and then to apply a 3D convolution neural network~(CNN) to extract features~(\citeauthor{3Dshapenets2015}, \citeyear{3Dshapenets2015}; \citeauthor{VoxNet2015}, \citeyear{VoxNet2015}). The volume representation retains relatively complete structural information of a point cloud, but this method is time-consuming and requires high computing memory costs. To this end, octree-based methods have been proposed  to reduce computational costs~(\citeauthor{O-CNN2017}, \citeyear{O-CNN2017}; \citeauthor{Octnet2017}, \citeyear{Octnet2017}). The second method is called multiview-based methods~(\citeauthor{MVCNN2015}, \citeyear{MVCNN2015}; \citeauthor{MHBN2018}, \citeyear{MHBN2018}; \citeauthor{GVCNN2018}, \citeyear{GVCNN2018}; \citeauthor{View-GCN2020}, \citeyear{View-GCN2020}). These methods project an unstructured 3D point cloud into pixel-based 2D maps~(e.g., LiDAR front view, bird's eye view~(BEV), and  the spherical map) and then use a well-established 2D-CNN to extract map features and fuse mapwise features from different view maps. The third method is to learn features directly from raw point clouds without any voxelization or projection. PointNet~(\citeauthor{PointNet2017}, \citeyear{PointNet2017}) was the first work to take raw point clouds as input. Specifically, PointNet extracts pointwise features with a multilayer perception~(MLP) layer and then uses a max-pooling function to generate global features. As a pioneering work, PointNet achieves state-of-the-art performance on the classification and segmentation task. However, this approach ignores local structural relationships between keypoints. Therefore, \cite{PointNetPlusPlus2017} proposed another hierarchical network, PointNet++, to obtain geometric structure information from the nearest neighbor of each point. The last is graph-based methods. Graph-based networks treat each point as graph vertices of a graph, and generate edges based on the nearest neighbors of each point. \cite{DGCNN2019}~proposed an unsupervised multitask algorithm DGCNN that constructs a local graph neural network and applies channelwise symmetric aggregation onto edges to connect the neighbors of each point. 
\subsection{Outlier Removal}
Due to the acquisition equipment and the environmental noise where the target object is located, the correspondences inevitably contain noise, which may cause some correspondences to become outliers. The existence of outliers may reduce the point cloud alignment accuracy, so it is necessary to take measures to filter out these outliers. The task of filtering outliers is also called inlier/outlier classification~(\citeauthor{3DRegNet2020}, \citeyear{3DRegNet2020}), where a correspondence is identified as whether an inlier or an  outlier according to a specific criterion. 
\par 
The traditional outlier removal methods include  RANdom SAmple Consensus~(RANSAC)~(\citeauthor{RANSAC1981}, \citeyear{RANSAC1981}) and its variants~(\citeauthor{GCRANSAC2018}, \citeyear{GCRANSAC2018}). The RANSAC method iteratively samples a small set of correspondences to ensure that the outliers  are filtered out as much as possible.  Other methods accomplish the task of outlier removal based on branch-and-bound~(BnB)~(\citeauthor{GoICP2016}, \citeyear{GoICP2016}), semidefinite programming~(SDP)~(\citeauthor{Sdrsac2019}, \citeyear{Sdrsac2019}; \citeauthor{Least_squares_registration2019}, \citeyear{Least_squares_registration2019}) or maximum clique schemes~(\citeauthor{maximumclique2019}, \citeyear{maximumclique2019}; \citeauthor{MaximalCliques2012}, \citeyear{MaximalCliques2012}). These methods generally require more sampling iterations and higher computational costs. However, the robustness of FGR~(\citeauthor{FastGlobalRegistration2016}, \citeyear{FastGlobalRegistration2016}) and TEASER~(\citeauthor{TEASER2021}, \citeyear{TEASER2021}) remains strong in the presence of high outlier rates. The main learning-based outlier filtering schemes are DGR~(\citeauthor{DeepGlobalRegistration2020}, \citeyear{DeepGlobalRegistration2020}) and 3DRegNet~(\citeauthor{3DRegNet2020}, \citeyear{3DRegNet2020}). The DGR algorithm uses a 6D convolutional network to classify the inliers and outliers, while the 3DRegNet algorithm uses multilayer ResNets as the classifier.

\subsection{Point Cloud Registration}
The iterative closet point~(ICP)~(\citeauthor{ICP1992}, \citeyear{ICP1992}), which alternately performs correspondence searching and least squares optimization to update the alignment state, is the best-known algorithm used for solving rigid registration problems. The literature~(\citeauthor{ReviewPointCloudReg2015}, \citeyear{ReviewPointCloudReg2015}; \citeauthor{EfficientVariantsofICP2001}, \citeyear{EfficientVariantsofICP2001}) summarized ICP and its variants in the last 20 years. The performance of the ICP algorithm is highly dependent on the accuracy of the initial estimated pose. However, the initial estimated information obtained from the odometers is not always reliable and may easily fall into local optima. To find an optimal transformation with ICP, \cite{GoICP2016} proposed the Go-ICP algorithm to determine the global optimal poses. It outperforms the ICP algorithm when point cloud registration requires the provision of a globally optimal solution. In addition, other algorithms attempt to apply convex relaxation~(\citeauthor{ConvexRelaxation2016}, \citeyear{ConvexRelaxation2016}), Riemannian optimization~(\citeauthor{SE-Sync2019}, \citeyear{SE-Sync2019}), graph-matching-based correspondence search method~(\citeauthor{Graph_matching_based_correspondence_search}, \citeyear{Graph_matching_based_correspondence_search}), and mixed-integer programming~(\citeauthor{MixedIntegerProgramming2020}, \citeyear{MixedIntegerProgramming2020}) to determine the global optimal pose, but these methods are computationally expensive and cannot meet the practical application requirements well. 

\begin{figure*}[hbt]
    \begin{subfigure}[c]{4.2cm}
        \begin{minipage}[t]{0.33\linewidth}
            \includegraphics[width=3.5\linewidth]{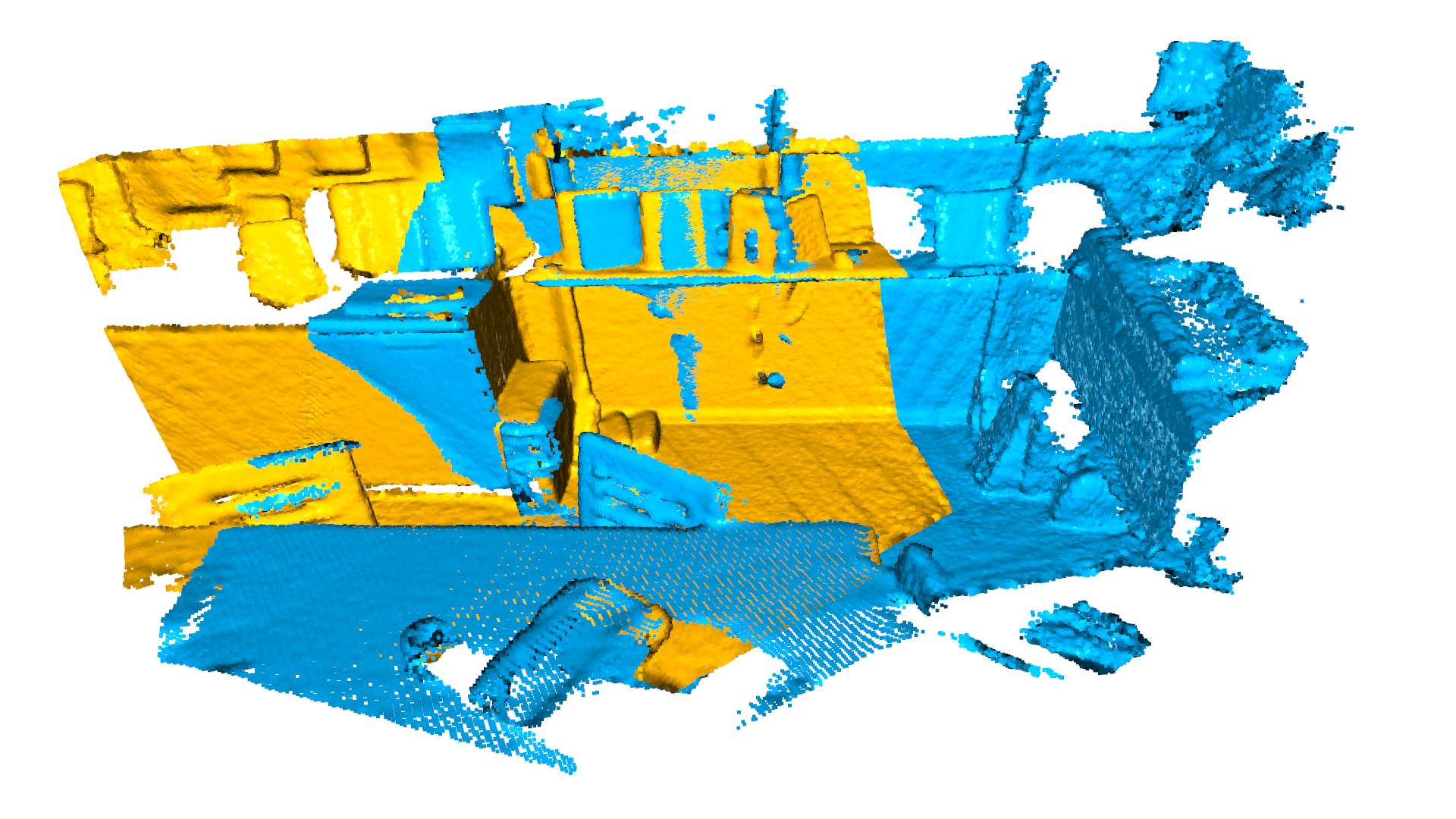}\vspace{0.3cm}
            \includegraphics[width=3.5\linewidth]{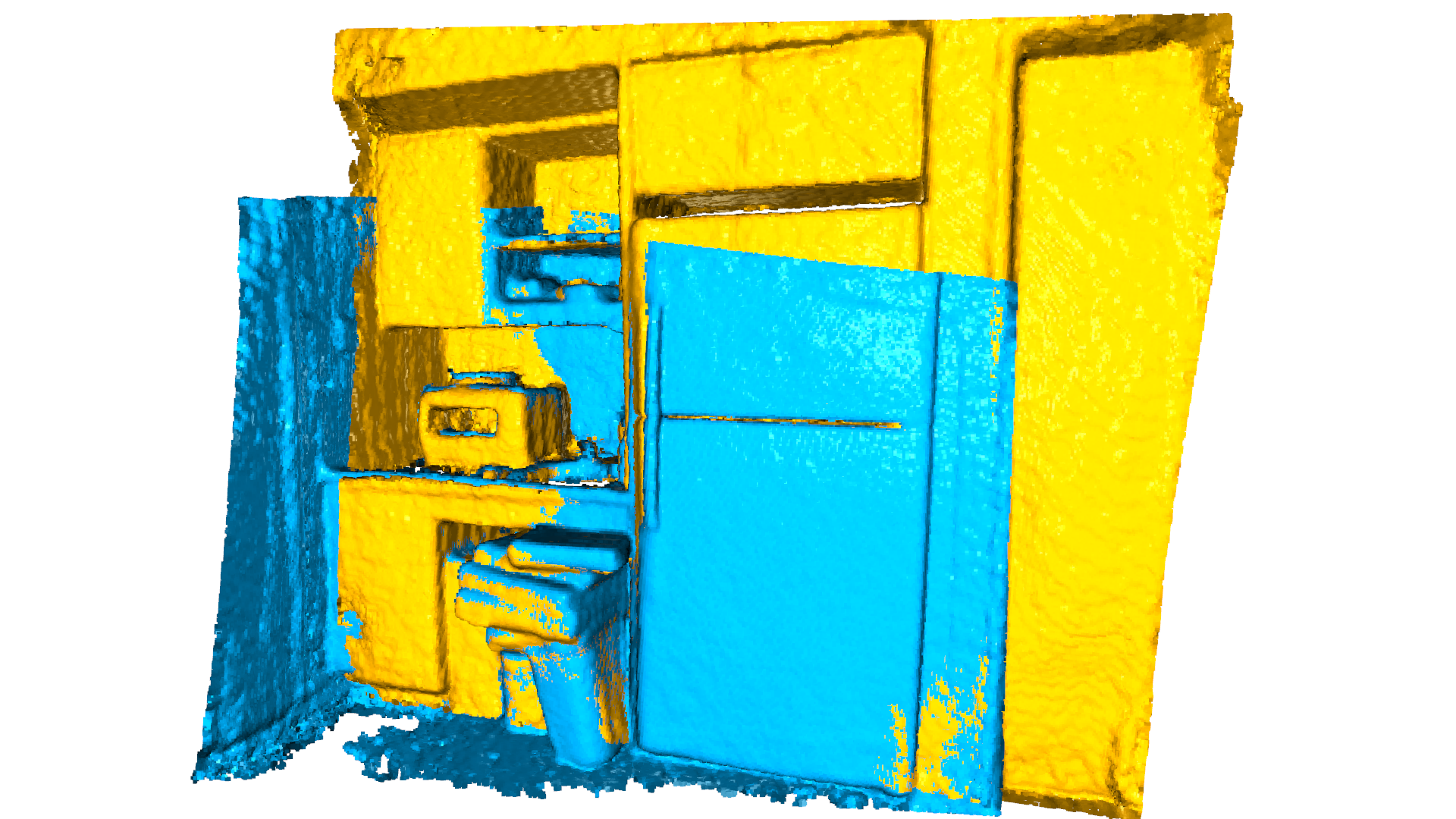}\vspace{0.5cm}
        \end{minipage}
        \subcaption{RANSAC (\citeauthor{RANSAC1981}, \citeyear{RANSAC1981})}
    \end{subfigure}
    \begin{subfigure}[c]{4.2cm}
        \begin{minipage}[t]{0.33\linewidth}
            \includegraphics[width=3.5\linewidth]{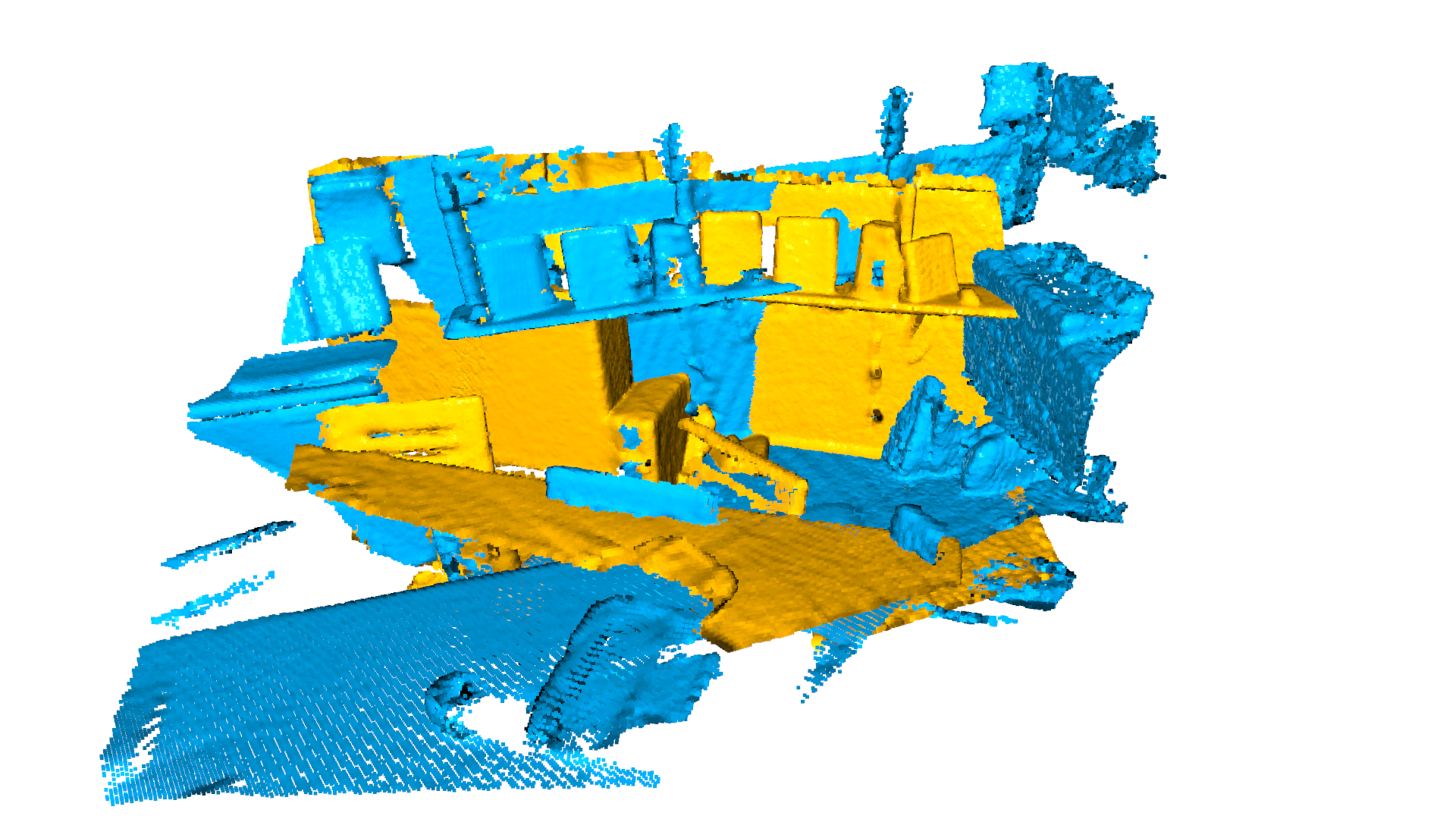}\vspace{0.3cm}
            \includegraphics[width=3.5\linewidth]{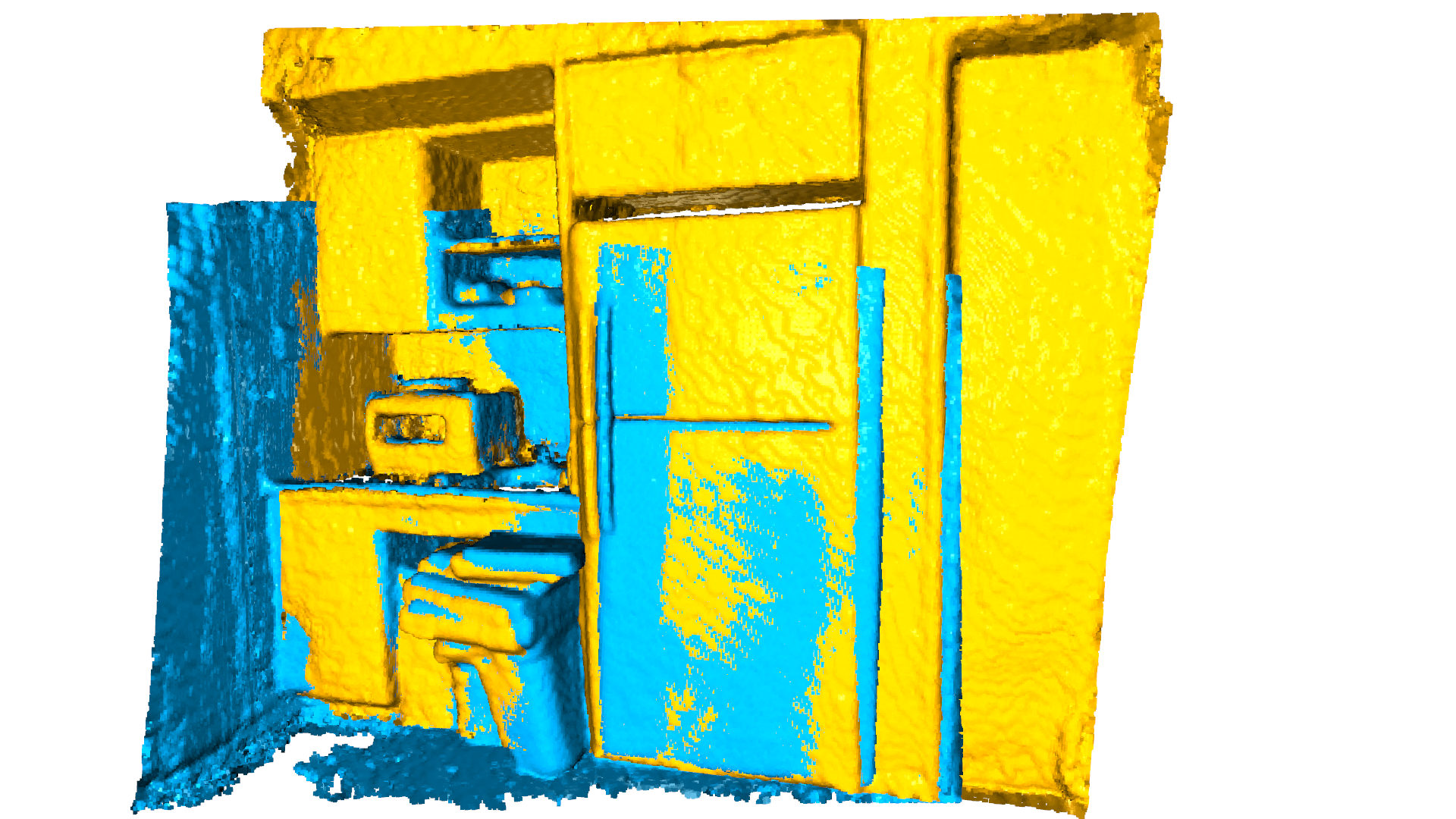}\vspace{0.3cm}
        \end{minipage}
        \subcaption{DGR (\citeauthor{DeepGlobalRegistration2020}, \citeyear{DeepGlobalRegistration2020})}
    \end{subfigure}
    \begin{subfigure}[c]{4.2cm}
        \begin{minipage}[t]{0.33\linewidth}
            \includegraphics[width=3.5\linewidth]{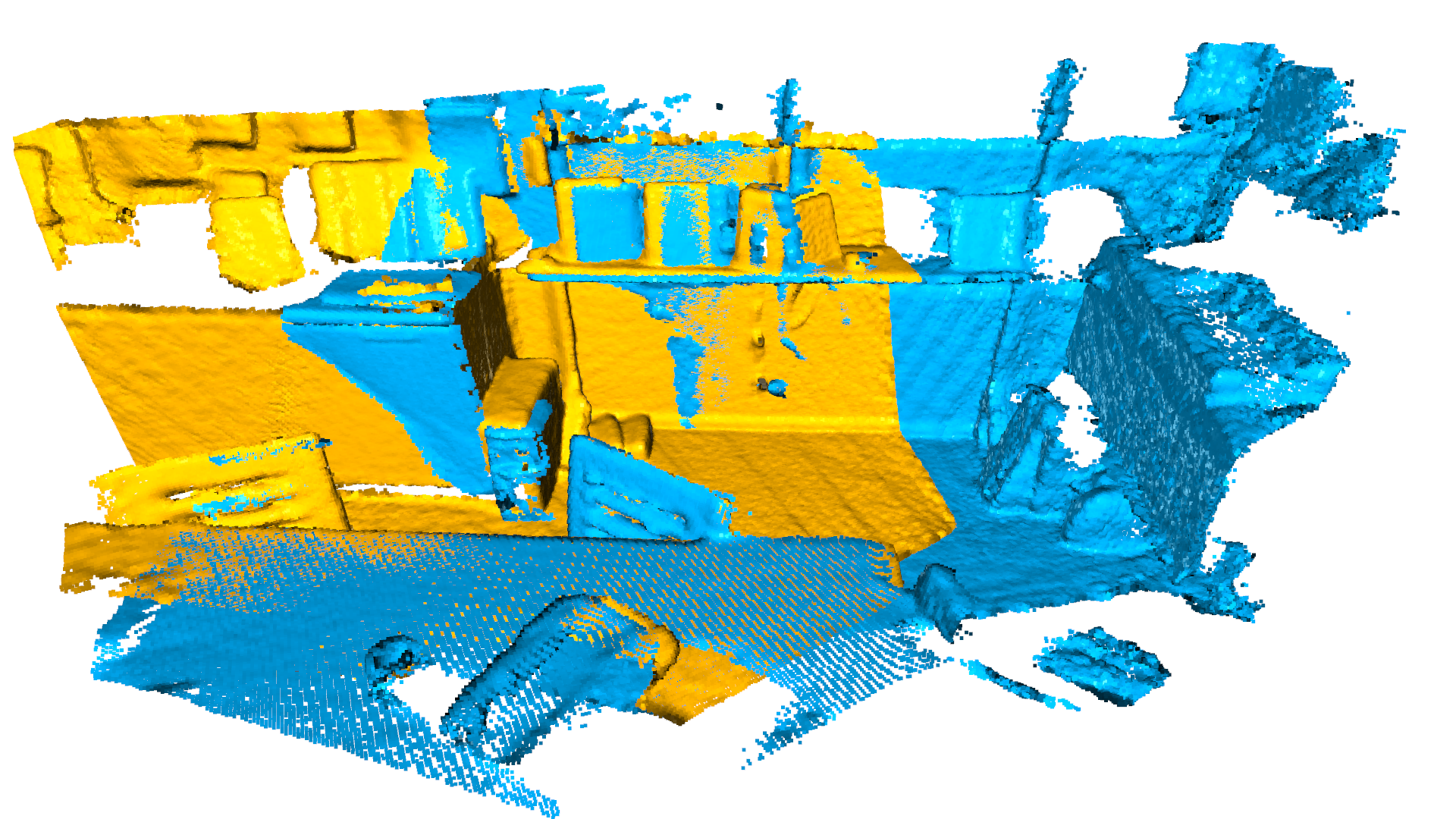}\vspace{0.3cm}
            \includegraphics[width=3.5\linewidth]{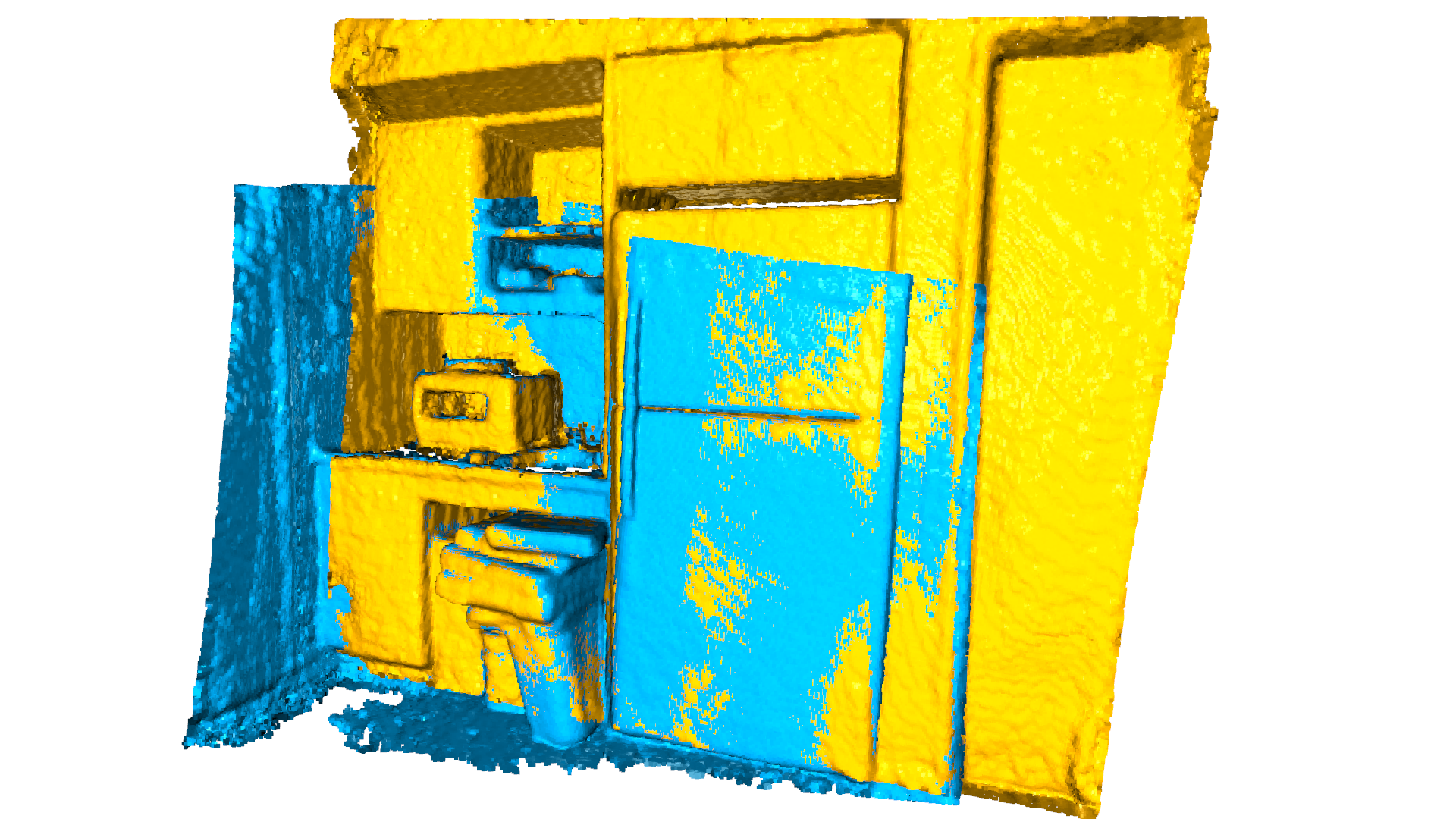}\vspace{0.3cm}
        \end{minipage}
        \subcaption{PointDSC (\citeauthor{PointDSC2021}, \citeyear{PointDSC2021})}
    \end{subfigure}
    \begin{subfigure}[c]{4.2cm}
        \begin{minipage}[t]{0.33\linewidth}
            \includegraphics[width=3.5\linewidth]{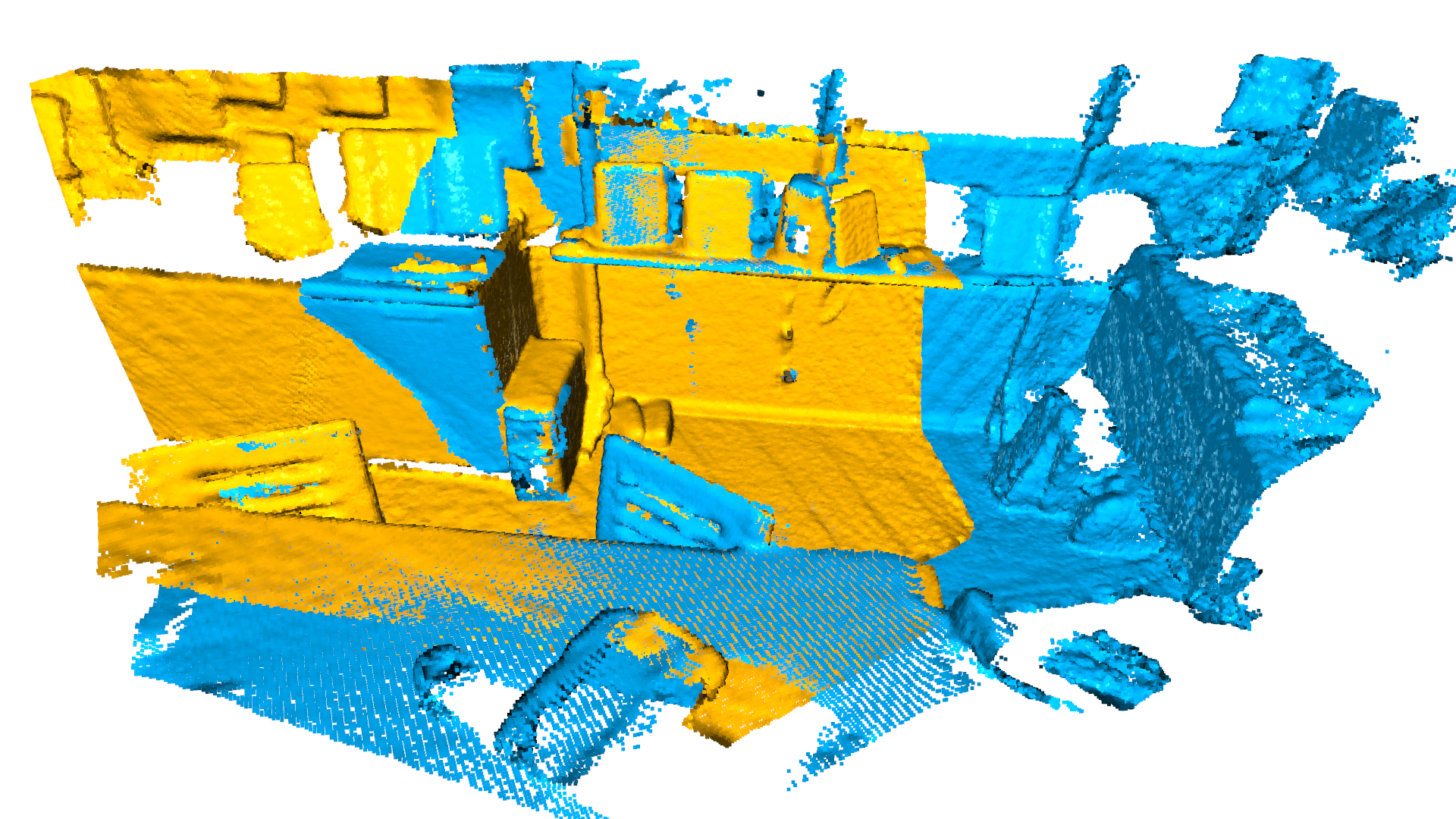}\vspace{0.3cm}
            \includegraphics[width=3.5\linewidth]{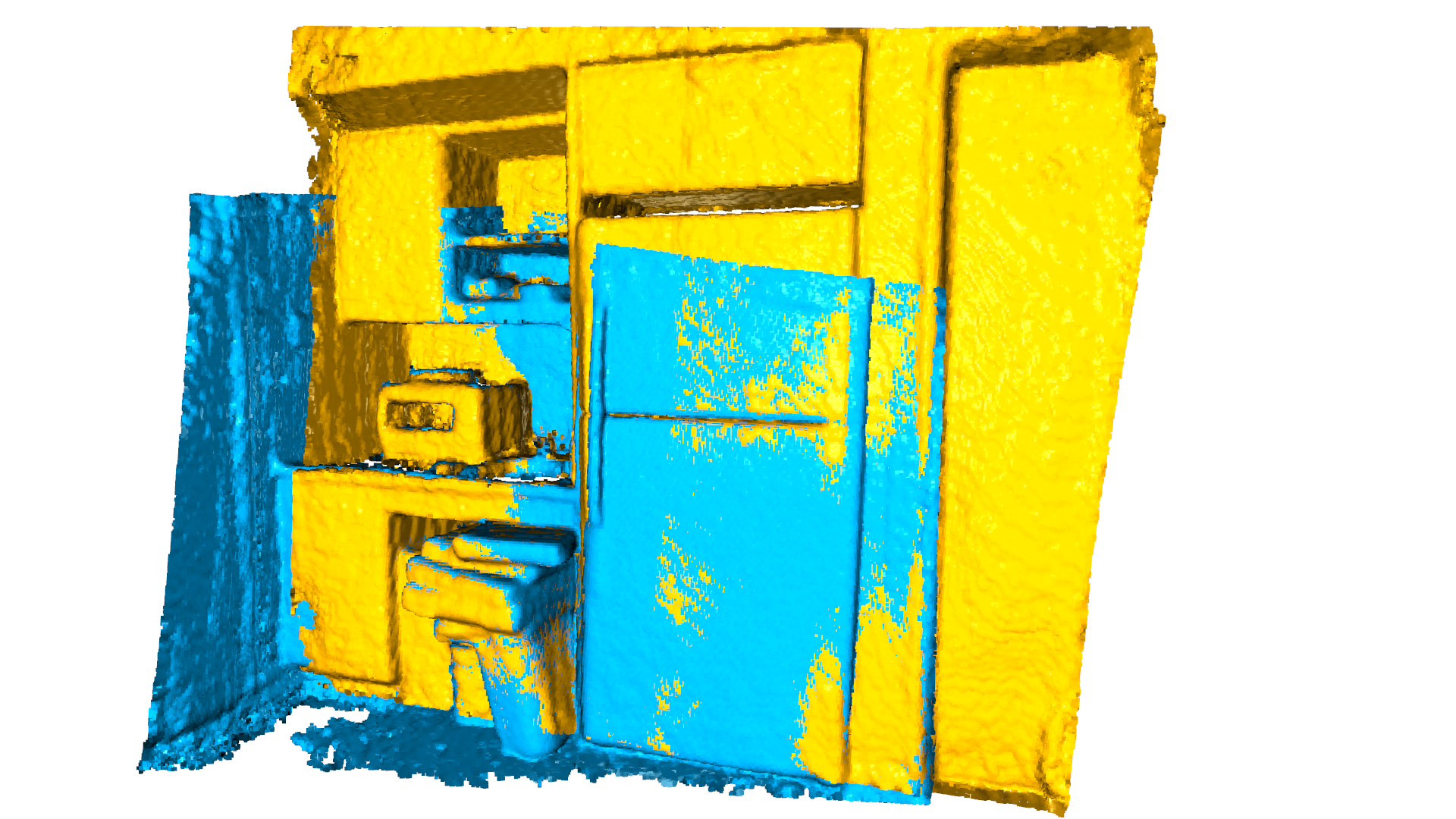}\vspace{0.3cm}
        \end{minipage}
        \subcaption{Ours}
    \end{subfigure}
    \caption{ Results of the transformation estimation that aligns two point clouds, ours  vs. the current state-of-the-art RANSAC, DGR and PointDSC. Our method achieves more satisfactory and perfect alignment than RANSAC, DGR and PointDSC.}
    \label{FigPairwise registration results}
\end{figure*}
\par 
In recent years, the application of deep learning has made great progress in point cloud registration. PointNetLK~(\citeauthor{PointNetLK2019}, \citeyear{PointNetLK2019}) combines global feature descriptors based on PointNet~(\citeauthor{PointNet2017}, \citeyear{PointNet2017}) and the Lucas/Kanade optimization algorithm~(\citeauthor{LucasKanade1981}, \citeyear{LucasKanade1981}) and then iteratively solves the relative rigid transformation.  DGR~(\citeauthor{DeepGlobalRegistration2020}, \citeyear{DeepGlobalRegistration2020}) uses a ConvNet to estimate the inlier likelihood of each correspondence and then applies a weighted Procrustes method to align point clouds. However, 3D spatial relations are omitted in these registration pipelines~(\citeauthor{DeepGlobalRegistration2020}, \citeyear{DeepGlobalRegistration2020}; \citeauthor{RANSAC1981}, \citeyear{RANSAC1981}). To focus on leveraging the spatial consistency in outlier rejection, \cite{PointDSC2021} proposes a spatial-consistency guided nonlocal module for geometric feature embedding of the correspondences and then uses a neural spectral matching~(NSM) module to compute the rigid transform for each seed. Different from the above networks, Predator~(\citeauthor{PREDATOR2021}, \citeyear{PREDATOR2021}) uses a parallel encode-decode structure and proposes a deep attention mechanism for the overlapping regions to exchange information about two unaligned point clouds.

%% file: includefiles/section03.tex
\section{Problem Formulation}\label{sec03Problem Formulation}
In general, point cloud fragments are obtained with the light detection and ranging~(LiDAR) scanners by receiving laser beams reflected by objects in the surrounding environment~(\citeauthor{Deep_structural_Information_fusion2022}, \citeyear{Deep_structural_Information_fusion2022}). The task to align two or more point clouds by estimating the relative transformation between them is named the point cloud registration or pose estimation. Given two point clouds:
\begin{align}
    P &= \{{{x}_{i}},i=1,2,\cdots ,M\}\\
    Q &= \{{{y}_{i}},i=1,2,\cdots ,N\}
\end{align}
where $P$ and $Q$ denote the source and target point clouds, respectively, and ${{x}_{i}}\in {{\mathbb{R}}^{3}}$ and ${{y}_{i}}\in {{\mathbb{R}}^{3}}$ are the 3D point coordinates of the source and target point cloud fragments. For ease of notation, we only describe the simplest case of point cloud registration, in which $M=N$ and $\{({{x}_{i}},{{y}_{i}})\}_{i=1}^{N}$. 
\par
The object of point cloud registration is to estimate the relatively rigid transformation that can correctly align two point clouds. We denote the rigid transformation as $[ {\mathbf{R}},{\mathbf{t}} ]$, which can be represented as follows: 
\begin{equation}
    [ {\mathbf{R}},{\mathbf{t}} ]=\text{arg}  \underset{\mathbf{R},\mathbf{t}}  {\mathop{\min }}\,\frac{1}{N}\sum\limits_{i}^{N}{{{w}_{i}}{{\left\| \mathbf{R}{{x}_{i}}+\mathbf{t}-{{y}_{i}} \right\|}^{2}}}
    \label{EqPointCloudReg}
\end{equation}
in which $\mathbf{R}\in SO(3)$ and $\mathbf{t}\in \mathbb{R}^{3}$ denotes the rotation matrix and the translation vector, $(x_i,y_i)$ is a pair of matched correspondence points, and ${w}_{i}$ indicates an inlier  likelihood for a certain correspondence $( {x}_{i},{y}_{i} )$.
\\
First, the  weighted centroids of  $P$ and $Q$  are defined as:
\begin{equation}
         \overline{x}=\frac{\sum\limits_{i=1}^{N}{{{w}_{i}}{{x}_{i}}}}{\sum\limits_{i=1}^{N}{{{w}_{i}}}}, \\
        \overline{y}=\frac{\sum\limits_{i=1}^{N}{{{w}_{i}}{{y}_{i}}}}{\sum\limits_{i=1}^{N}{{{w}_{i}}}}      
\end{equation}
Then, the next step is to compute cross-covariance matrix $\mathbf{H}$:
\begin{equation}
    \mathbf{H}=\sum\limits_{i=1}^{N}{{{w}_{i}}({{x}_{i}}-\overline{x})}{{({{y}_{i}}-\overline{y})}^{T}}
    \label{EqCross-CovarianceMatrix}
\end{equation}
Last, we need to use singular value decomposition(SVD) method to decompose $\mathbf{H}$:
\begin{equation}
    \left[ \mathbf{U,S,V} \right] =\text{SVD}(\mathbf{H})
\end{equation}
Eq.~\ref{EqPointCloudReg}  gives a closed-form solution to the rigid transformation by minimizing the mean-square error~(MSE) function:
\begin{equation}
        E_{1}=\frac{1}{N}\sum\limits_{i}^{N}{{{w}_{i}}{{\left\| \mathbf{R}{{x}_{i}}+\mathbf{t}-{{y}_{i}} \right\|}^{\text{2}}}}
\end{equation}
The rigid transformation $[{\mathbf{R}},{\mathbf{t}}]$ can be obtained as follows:
\begin{align}
    {\mathbf{R}}& =\mathbf{V}diag(1,1,\cdots ,\mathrm{det}(\mathbf{V}{\mathbf{U}^\mathrm{T}})){\mathbf{U}^\mathrm{T}}\\
    {\mathbf{t}}& =-{\mathbf{R}}\overline{x}+\overline{y}
\end{align}
where det($\cdot$) denotes the determinant.

%% file: includefiles/section04.tex
\section{Deep Feature Consistency}\label{sec03NetworkDesign}
\begin{figure*}[hbt]
    \includegraphics[scale=0.8]{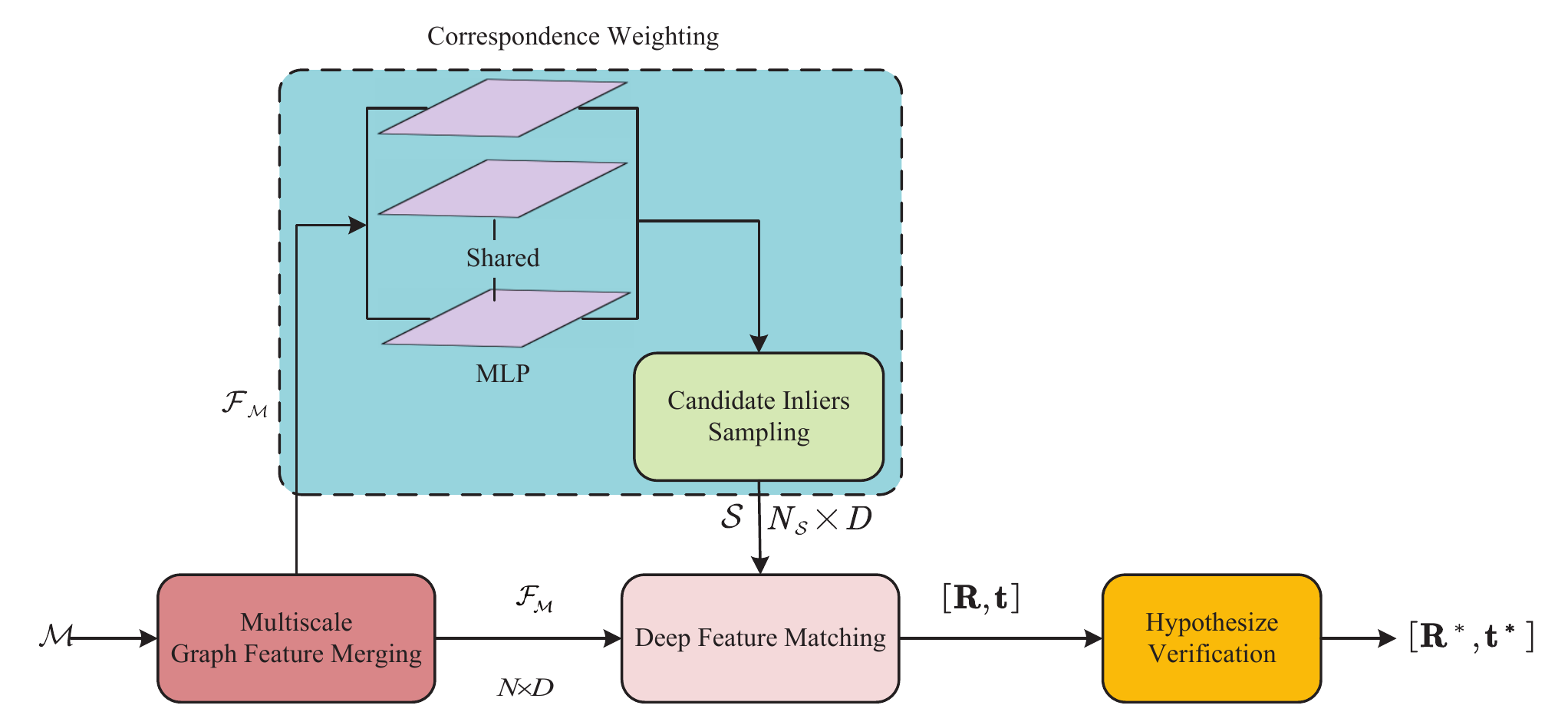}
    \caption{Overall framework of the proposed DFC pipeline. It takes the correspondence set $\mathcal{M}$ as input, outputs the best rigid transformation and classifies the correspondences as inliers and outliers.}
    \label{FigPipeline}
\end{figure*}
Having  demonstrated preliminaries about point cloud registration, we are now equipped to present the architecture of the proposed deep feature consistency, abbreviated as DFC. The overall architecture of  our DFC pipeline is shown in Fig.~\ref{FigPipeline}~with three modules for feature embedding, correspondence weighting and deep feature matching. The goal of the proposed DFC method for registration is to provide an excellent end-to-end solution for correctly classifying correspondences into outliers/inliers and estimating the relative transformation between two unaligned point clouds in the absence of initial transformation prediction.
\par 
In summary, our pairwise registration pipeline begins with  embedding deep features of the putative correspondences into high dimensional space. Then, we apply the correspondence weighting module to predict the veracity of each correspondence and sample small correspondences to form multiple candidate inlier subsets. Last, we use a deep feature matching module to estimate the rigid transformation for each subset and determine the best rigid transformation among these transformations.
\par 
It is necessary to take steps to preprocess the point cloud before feeding the original point cloud data into our pipeline. Preprocessing tasks can be divided into two steps: pointwise feature encoding and data augmentation. Here we only discuss the pointwise feature encoding task, and the data augmentation details  are described in Sec.~\ref{sec05Experiments}. Similar to KPconv~(\citeauthor{Kpconv2019}, \citeyear{Kpconv2019}), DGR~(\citeauthor{DeepGlobalRegistration2020}, \citeyear{DeepGlobalRegistration2020}) and PointDSC~(\citeauthor{PointDSC2021}, \citeyear{PointDSC2021}), first, we  use a voxelized grid filter to downsample the original point clouds $P$ and $Q$ in the feature encoding process, which can ensure $P$ and $Q$ have a reasonable point density and thus immensely reduce the computational costs. To be able to describe geometric semantic information in the form of vectors in the feature space, fully convolutional geometric features~(FCGF)~(\citeauthor{FullyConvolutionalGeometricFeatures2019}, \citeyear{FullyConvolutionalGeometricFeatures2019}) descriptors are used to extract pointwise features in the point cloud. FCGF descriptors can widely grasp spatial semantic information and can be easily extended to large-scale scenarios, which contributes to achieving ideal speedup. Then, we choose $N$ points with FCGF features from  the source point cloud by random sampling. The target point cloud also needs to suffer the same sampling processing. On the one hand, point sampling  can filter out the points with insignificant features to alleviate the influence of singular points on the registration effect; on the other hand, it can effectively reduce the computational costs and the registration time. The following notation is used throughout the paper. We can obtain the  features of  $N$ points from the two original point clouds: 
\begin{align}
    {{\mathcal{F}}_{x}}&=\{{\boldsymbol{f}_{{{x}_{1}}}} ,{\boldsymbol{f}_{{{x}_{2}}}} ,\cdots ,{\boldsymbol{f}_{{{x}_{N}}}}\}\\
    {{\mathcal{F}}_{y}}&=\{{\boldsymbol{f}_{{{y}_{1}}}} ,{\boldsymbol{f}_{{{y}_{2}}}} ,\cdots ,{\boldsymbol{f}_{{{y}_{N}}}}\}
\end{align}

\subsection{Feature Embedding}
The input of other state-of-the-art learning-based registration methods such as PointNetLK~(\citeauthor{PointNetLK2019}, \citeyear{PointNetLK2019}), DCP~(\citeauthor{DeepClosetPoint2019}, \citeyear{DeepClosetPoint2019}) and DeepVCP~(\citeauthor{DeepVCP2019}, \citeyear{DeepVCP2019}) only includes the coordinates of the key points located in the source and target point clouds, i.e, these methods independently extract the key-point features, thus ignoring the internal relation of the correspondences. In contrast to these methods, the input of our DFC method is not only the key points, but also the correspondence set $\mathcal{M}$:
\begin{equation}
    \mathcal{M} =\{({x_i},\mathrm{arg} \underset{{y_j}}{\min}\,\left\| \boldsymbol{f}_{x_i}-\boldsymbol{f}_{y_j} \right\| ^2)\left| i,j\in [1,2,\cdots ,N] \right\} 
\end{equation}
where $N$ denotes the number of correspondences. A correspondence $\left( {{x}_{i}}, {{y}_{j}} \right)$ can be represented as a specific point in 6-dimensional space $[ x_{i}^{\mathrm{T}}, y_{j}^{\mathrm{T}} ]^{\mathrm{T}}\in\mathbb{R}^6$~(\citeauthor{DeepGlobalRegistration2020}, \citeyear{DeepGlobalRegistration2020}). Therefore, it is out of question that we can consider embedding features of all correspondences along with the idea of pointwise feature extraction.
\begin{figure}[hbt]
    \centering
    \includegraphics[width=8cm]{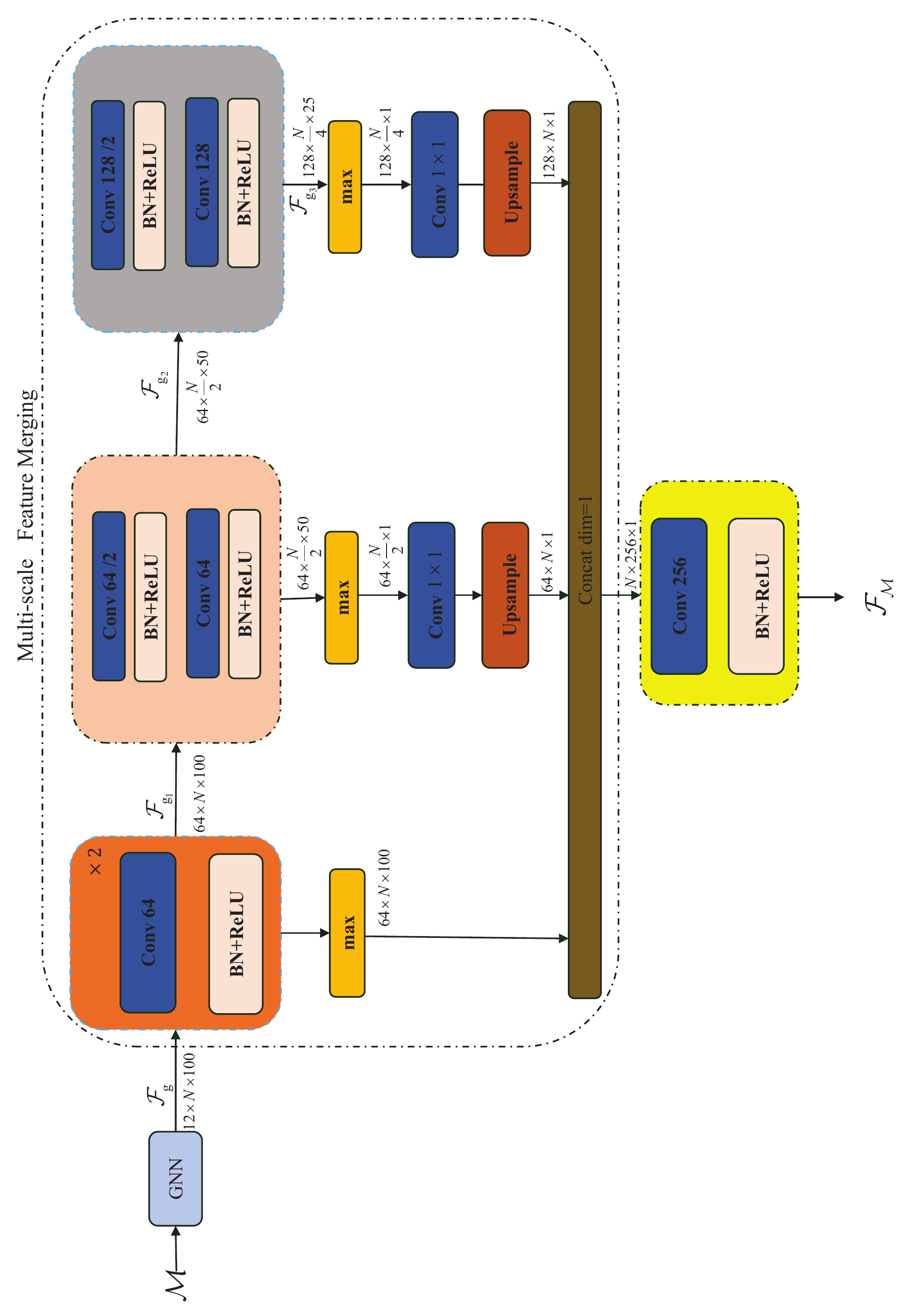}
    \caption{Multiscale graph feature merging(GFM) network architecture for feature embedding. It consists of two blocks: graph neural network(GNN) and multiscale feature merging(FM). The first GNN layer is used to extract the graph features of the correspondences, while a multiscale FM block is designed to grasp and fuse features at different scales. Best viewed on the screen.}
    \label{FigMulti-ScaleGFM}
\end{figure}
\par 
The process of generating the correspondence set can filter out the noise at a certain rate~(\citeauthor{FastGlobalRegistration2016}, \citeyear{FastGlobalRegistration2016}), thus effectively improving registration performance. It has been proven that  fusing feature maps from different scales can significantly improve the performance on a set of computer vision tasks~(\citeauthor{DeeplySupervisedSalientObjectDetection2017}, \citeyear{DeeplySupervisedSalientObjectDetection2017}; \citeauthor{FeaturePyramidNetworksforObjectDetection2017}, \citeyear{FeaturePyramidNetworksforObjectDetection2017}; \citeauthor{MultiLevelFeatureAggregation2021}, \citeyear{MultiLevelFeatureAggregation2021}; \citeauthor{FullyConvolutionalGeometricFeatures2019}, \citeyear{FullyConvolutionalGeometricFeatures2019}; \citeauthor{Siamrpn++2019}, \citeyear{Siamrpn++2019}), so we follow the idea of these excellent  methods to present multiscale graph feature merging~(GFM) network, which is designed for embedding features of the putative correspondences and compensating for the disadvantage of FCGF feature descriptors without geometric information of correspondences. The overall structure of the multiscale GFM network is shown in Fig.~\ref{FigMulti-ScaleGFM}, which consists of the graph neural network~(GNN)~(\citeauthor{DGCNN2019}, \citeyear{DGCNN2019}) and multiscale feature merging~(FM) block. 
\par
\textbf{Graph Neural Network.}
Let  $\boldsymbol{m}_i$ and $\boldsymbol{m}_j$ denote a pair of adjacent correspondences, and $\left( i,j \right) \in \varepsilon $ the graph edge between  $\boldsymbol{m}_i$ and $\boldsymbol{m}_j$. The correspondence $\boldsymbol{m}_i$ is used as the input of the GNN layer to obtain its graph feature representation using the $k$-nearest neighbor($k$-NN) in Euclidean space:
\begin{equation}
    \mathcal{F} _{g_{m_i}}=h_{\theta}\left( \boldsymbol{m}_i,\boldsymbol{m}_j \right) =\bar{h}_{\theta}\left( \boldsymbol{m}_i,\boldsymbol{m}_j-\boldsymbol{m}_i \right) 
\end{equation}
where the size of $\mathcal{F} _{g_{m_i}}$ is $12\times 1 \times 100$ and $\bar{h}_{\theta}\left( \cdot \right)$ denotes a nonlinear function with a series of learnable parameters $\theta$. 
\par 
By performing such steps for each correspondence among $\mathcal{M}$ in parallel, the final GNN features of all correspondences can be gained as ${\mathcal{F}}_{g}$ with the size of $12\times N \times 100$.
\par
\textbf{Multiscale Feature Merging.}
The multiscale FM block consists of three scale layers, each of which consists of two blocks of a $1 \times 1$ convolution function followed by batchnormalization~(BN)~(\citeauthor{BatchNormalization2015}, \citeyear{BatchNormalization2015}) and ReLU activation~(\citeauthor{ReLU2010}, \citeyear{ReLU2010}). The outputs ${\mathcal{F}}_{g}$ of GNN are then fed to the multiscale FM block. For the sake of description, let ~${\mathcal{F}}_{g_{1}}, {\mathcal{F}}_{g_{2}}$ and ${\mathcal{F}}_{g_{3}}$~ define the different output graph features of the three scale layers, respectively. The spatial size between the last adjacent scale layers is reduced by half with stride 2, and the channel number between the first and second scale layers is always maintained at 64 while it becomes twice as large as 128 under the third scale layer. To smoothly merge the output graph features of the three scale layers, the feature maps ${\mathcal{F}}_{g_{2}}$ and ${\mathcal{F}}_{g_{3}}$ are upsampled to the same spatial size as ${\mathcal{F}}_{g_{1}}$ is. Finally, the graph features from the  three scale layers mentioned above are merged and fed into another convolution with a $1 \times 1$ kernel and 256 channels, followed by BN and ReLU layer to obtain the final features ${{\mathcal{F}}_{\mathcal{M}}}$ with the size of $N \times D$, where $D$ means the feature embedding dims and is set to $256$. 
\par 
Let $\mathcal{F} _{\mathcal{M}}=\left\{ \boldsymbol{f}_1,\boldsymbol{f}_2,\cdots ,\boldsymbol{f}_N \right\}$ ~denote the final features of ${\mathcal{M}}$ and $\boldsymbol{f}_i$ represent a feature representation for the $i$-th correspondence $\left( {{x}_{i}}, {{y}_{i}} \right)$. The feature representations ${{\mathcal{F}}_{\mathcal{M}}}$ contain both high-level semantic information and low-level detail, which allows us to set a better balance between invariance and discriminability~(\citeauthor{MultiLevelFeatureAggregation2021}, \citeyear{MultiLevelFeatureAggregation2021}). 
\subsection{Correspondence Weighting}
\label{SecCorrespondenceWeighting}
A putative correspondence with obvious features should ideally be assigned with a higher weight. Similar to DeepVCP~(\citeauthor{DeepVCP2019}, \citeyear{DeepVCP2019}), AdaLAM~(\citeauthor{Adalam2020}, \citeyear{Adalam2020}) and PointDSC~(\citeauthor{PointDSC2021}, \citeyear{PointDSC2021}), we design a correspondence weighting module to sample small correspondences with high confidence as a candidate inlier set, and then search for adjacent correspondences for each candidate inlier in the feature space to form multiple inlier subsets that will replace ${\mathcal{M}}$ as the input of the subsequent deep feature matching module. Compared with the whole correspondence set ${\mathcal{M}}$, the candidate inliers have much more probabilities to be good candidates, which can better guarantee the registration success ratio.
\par 
As shown in Fig.~\ref{FigPipeline}, First, we  decide to apply an MLP to estimate a likelihood for each correspondence among $\mathcal{M}$ based on the features ${{\mathcal{F}}_{\mathcal{M}}}$ extracted in the previous multiscale GFM network. This estimation can be viewed as likely if a correspondence paired is an inlier. The higher the confidence predicted by the MLP, the more likely the counterpart correspondence  may be a good candidate to be an inlier.
As the core of the correspondence weighting module, the MLP layer is composed of the three fully connected layers, where the first two layers consist of a convolution function followed by ReLU, and the last layer only consists of the convolution function, excluding ReLU activation. Then in the sampling step, we select a series of correspondences $\mathcal{S} \left( \mathcal{S} \subseteq \mathcal{M} \right)$~based on the confidence ranking in descending order by direct one-shot sampling, unlike the iterative sampling optimization of the RANSAC method. Here, we finally determine that the total sampling number $N_{\mathcal{S}}$ of $\mathcal{S}$ is set to $200$. These selected correspondences have many unique characteristics that may enhance the registration performance, including high confidence and wide distribution. These selected correspondences are referred to as candidate inliers for the sake of distinction. 
\par 
In the process of sampling candidate inliers, the correspondences among ${\mathcal{M}}$ with low confidence are identified as latent outliers, which completes the task of coarse filtering outliers. In the subsequent Sec.~\ref{Sec5.2PairwiseRegistration}, we conduct experiments to explore the effectiveness of outlier removal and how it affects the alignment accuracy.
\subsection{Deep Feature Matching}
\begin{figure}[h]
    \centering
    \includegraphics[width=8cm]{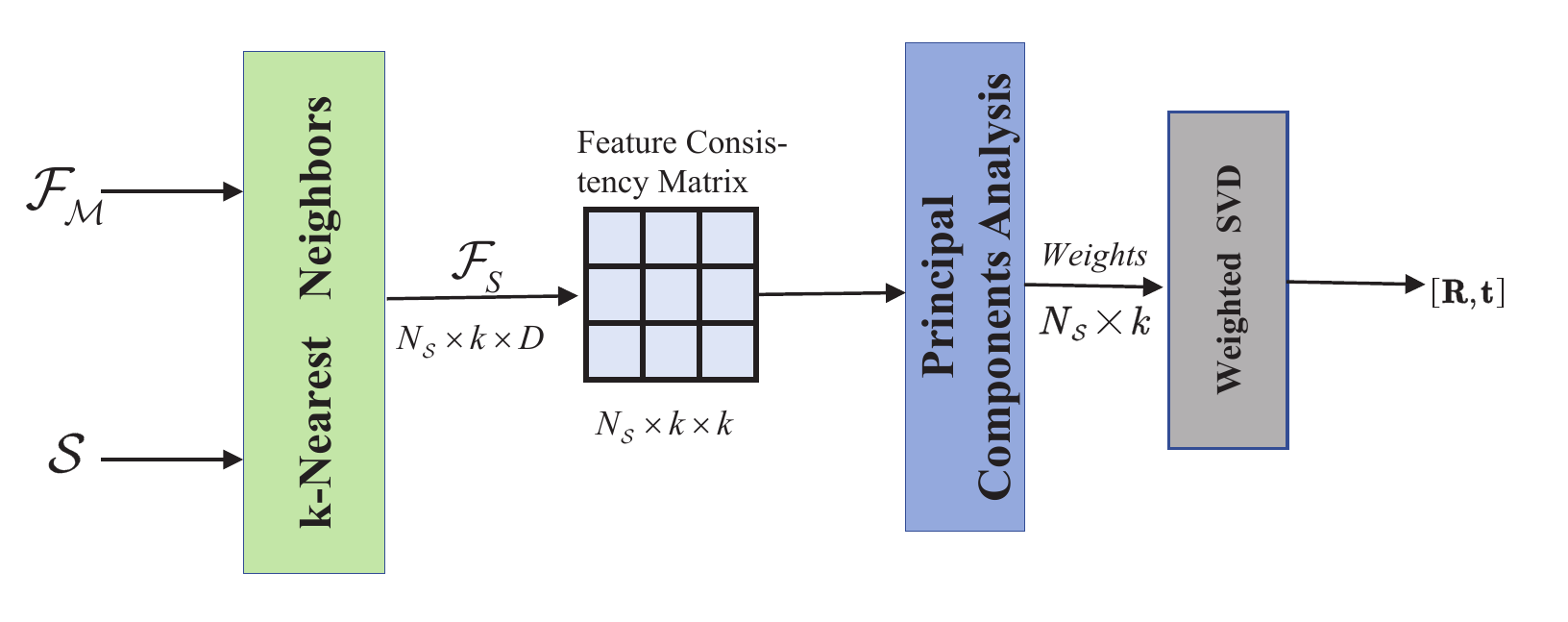}
    \caption{The detailed steps of the deep feature matching module used for estimating a rigid transformation for each candidate inlier subset.}
    \label{FigDeepFeatureConsistency}
\end{figure}
Obtaining a series of candidate inliers $\mathcal{S}$ from the previous correspondence weighting module, we construct a series of candidate inlier subsets ${\mathcal{C}}$ for each latent inlier in the feature space using the $k$-NN method, where $\left| \mathcal{C} \right|=k$ and $k$ takes the value of $40$. As shown in Fig.~\ref{FigDeepFeatureConsistency}, the overall procedure of the deep feature matching module is divided into two steps: constructing a deep feature consistency matrix and estimating the rigid transformation by the weighted SVD method. First, we need to establish a deep feature consistency matrix to estimate the inlier probability for each candidate inlier subset among $\mathcal{C}$ and then to apply the weighted SVD method to solve the corresponding rigid transformation. 
\par 
The core of the deep feature matching module is  constructing  the feature consistency matrix $\mathbf{M}$ that derived from the literature~(\citeauthor{PairwiseConstraints2005}, \citeyear{PairwiseConstraints2005}) and its elements can be calculated according to Eq.~\ref{EqFeatureConsistencyMatrix}:
\begin{equation}
    e_{ij}=\left[ 1-\frac{1}{\sigma^{2}}\left\| {{\overline{\boldsymbol{f}}}_{i}}-{{\overline{\boldsymbol{f}}}_{j}} \right\|^{2} \right]
    \label{EqFeatureConsistencyMatrix}
\end{equation}
where ${{\overline{\boldsymbol{f}}}_{i}}$ and ${{\overline{\boldsymbol{f}}}_{j}}$ are the L2-normalized feature vectors of $\boldsymbol{f}_i$ and $\boldsymbol{f}_j$, respectively, and $\sigma$ is a balanced parameter to control the sensitivity to the feature difference~(\citeauthor{PointDSC2021}, \citeyear{PointDSC2021}). 
\par 
The calculation to solve $e_{ij}$ makes $\mathbf{M}$ nonnegative, serving as a role to ensure the consistency between the correspondence features. After computing $\mathbf{M}$ by Eq.~\ref{EqFeatureConsistencyMatrix}, the PCA method is then used to estimate its principal vectors. Here we denote the principal vectors as $\boldsymbol{w}=\left\{ w_1,w_2,\cdots ,w_k \right\}$. In Sec.~\ref{sec5AblationStudies}, we will give an ablation study on another method named eigenvalues to calculate principal vectors. The principal component  $\boldsymbol{w}$ can be considered as the inlier probability corresponding to each candidate inlier subset. Once we compute the inlier probability, the rigid transformation can be completed with Eq.~\ref{EqPointCloudReg}. Similarly, by performing the same steps for each candidate inlier subset in parallel, the rigid transformations $[{\mathbf{R}},{\mathbf{t}}]$ corresponding to each subset can be obtained simultaneously. 
\subsection{Hypothesis Verification}
The final step of the proposed DFC method is the same as the literature~(\citeauthor{PointDSC2021}, \citeyear{PointDSC2021}), i.e., to determine the optimal transformation $[ \mathbf{R}^*,\mathbf{t}^* ]$ among a series of rigid transformations $[ {\mathbf{R}},{\mathbf{t}} ]$ generated by the deep feature matching module according to a certain rule that the number of inliers computed by each transformation maximizes. The process of choosing the optimal transformation can be accomplished by maximizing the following objective function:
\begin{equation}
    {{E}_{2}}=\underset{[{\mathbf{R}},{\mathbf{t}}] }{\mathop{max}}\,\sum\limits_{i=1}^{k}{\left[\!\left[ \left\| {\mathbf{R}}{{x}_{i}}+{\mathbf{t}}-{{y}_{i}} \right\|<\tau  \right]\!\right]}
    \label{EqHypothesisVerification}
\end{equation}
where $\tau$ denotes the given inlier threshold and $ \llbracket \cdot \rrbracket $ denotes the Iverson bracket. When $\| {\mathbf{R}}{{x}_{i}}+{\mathbf{t}}-{{y}_{i}} \|$ is less than the given threshold $\tau$,  the correspondence $ \left( {x}_{i},{y}_{i} \right)$ is considered an inlier~(labeled as one), and the total number of inliers will be increased by one; otherwise,  $ \left( {x}_{i},{y}_{i} \right)$ will be identified as an incorrect correspondence~(labeled as zero). 
\subsection{Loss Function}
The loss function of the proposed DFC method consists of two independent loss terms called classification loss and transformation loss.
\par 
\textbf{Classification Loss.} The classification loss is a common metric to evaluate incorrect correspondences using binary cross entropy~(BCE)~(\citeauthor{3DRegNet2020}, \citeyear{3DRegNet2020}; \citeauthor{DeepGlobalRegistration2020}, \citeyear{DeepGlobalRegistration2020}):
\begin{equation}
    {\mathcal{L}_{c}}=\text{BCE}\left( \boldsymbol{c}, \boldsymbol{l}\right)
\end{equation}
where $\boldsymbol{c}$ is the confidence of the correspondence set computed by the MLP mentioned in Sec. \ref{SecCorrespondenceWeighting}.  $\boldsymbol{l}=\left\{ l_1, l_2, \cdots, l_N \right\}$, where $l_i$ ~(equal to one or zero) is the ground-truth label, which indicates  the $i$-th point correspondence is whether an inlier or outlier. 
\par
\textbf{Transformation Loss.}
The transformation loss is commonly used to assess the agreement between the ground-truth rigid transformation $[{\mathbf{R}^{g}},{\mathbf{t}^{g}}]$ and estimated rigid transformation $[{\mathbf{R}^{*}},{\mathbf{t}^{*}}]$:
\begin{equation}
    {{\mathcal{L}}_{t}}\text{=}{{\left\| {{({\mathbf{R}^{*}})}^{\mathrm{T}}}{\mathbf{R}^{g}}-I \right\|}^{2}}+{{\left\| {\mathbf{t}^{*}}-{\mathbf{t}^{g}} \right\|}^{2}}
\end{equation}
\par 
The total loss is a weighted sum of the above loss functions:
\begin{equation}
    \mathcal{L}={{\mathcal{L}}_{c}}+\lambda {\mathcal{L}_{t}}
\end{equation}
where $\lambda$ is a hyperparameter that can be  manually set to balance these two losses. 

%% file: includefiles/section05.tex
\section{Experiments}\label{sec05Experiments}
In this section, we analyze the robustness and generalization of the proposed DFC method in indoor, outdoor and multiway registration scenarios. As the name suggests, pairwise registration means estimating the rigid transformation between two point cloud scans as shown in Fig.~\ref{FigPairwise registration results}, and multiway registration produces a final global reconstruction map and pose estimation for all point cloud fragments. Pairwise registration plays an extremely important role in the  multiway registration task. Before performing multiway registration, we need to use a pairwise registration method to estimate the initial poses and then obtain the optimal poses with robust pose graph optimization.
\par 
For indoor alignment scenarios, we choose the 3DMatch dataset~(\citeauthor{3DMatch2017}, \citeyear{3DMatch2017})  to evaluate the performance of our method, where the scans are composed of 3D point clouds from different real-world scenes, and these point cloud scans also contain ground-truth transformations computed by the RGB-D reconstruction system.  To verify the generalization ability of our pipeline across different datasets, we conduct another cross-dataset experiment on the augmented ICL-NUIM~(\citeauthor{RobustReconstructionofIndoorsScenes2015}, \citeyear{RobustReconstructionofIndoorsScenes2015}; \citeauthor{ICL-NUIM2014}, \citeyear{ICL-NUIM2014}) to quantify the average trajectory error~(ATE). In addition, we use KITTI odometry~(\citeauthor{KITTIDataset2012}, \citeyear{KITTIDataset2012}) as a benchmark dataset for large outdoor alignment scenarios. However, there is no clear official division labels for train/val/test splits, so we are determined to divide the KITTI odometry benchmark into train/val/test sets following FCGF~(\citeauthor{FullyConvolutionalGeometricFeatures2019}, \citeyear{FullyConvolutionalGeometricFeatures2019}).
\par 
During training, we apply Gaussian noise with a standard deviation of 0.03, random rotations $\in [0^{\degree},360^{\degree})$ around a random axis. All experiments are performed on the platform with one single NVIDIA RTX 2080Ti graphics card and Intel Xeon E5-2630 v3 CPU. We construct our model in PyTorch, train it on the previous platform for 100 epochs and set the batch size to 8.
\subsection{Pairwise Registration}\label{Sec5.2PairwiseRegistration}
The scheme of the train/test splits is the same as (\citeauthor{FullyConvolutionalGeometricFeatures2019}, \citeyear{FullyConvolutionalGeometricFeatures2019}; \citeauthor{DeepGlobalRegistration2020}, \citeyear{DeepGlobalRegistration2020}; \citeauthor{PointDSC2021}, \citeyear{PointDSC2021}), i.e., 54 scenes among the 3DMatch dataset are used for training and validation, and the remaining  8 scenes are used  for testing. The hyperparameter $\lambda$ and the inlier threshold $\tau$ are set to $10^{-2}$ and $10cm$, respectively. A voxelized $5cm$ grid is first used to downsample the point cloud data, and then the FCGF descriptors are applied to extract the pointwise features to prepare to construct the input correspondences. 
\par
\begin{figure}[hbt]
    \begin{subfigure}[c]{3.9cm}
        \includegraphics[width=5cm]{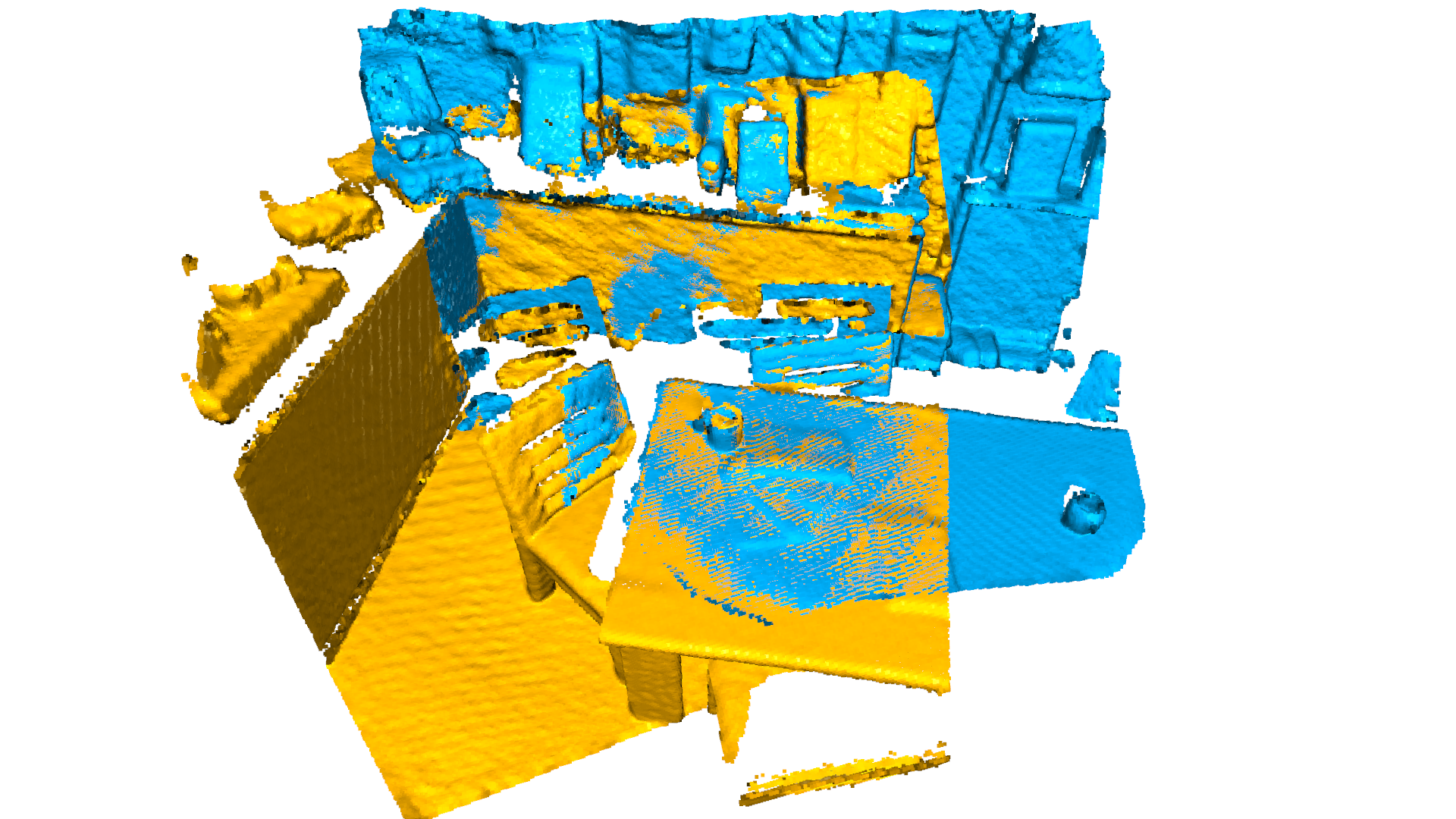}
        \subcaption{Kitchen}
    \end{subfigure}
    \quad
    \begin{subfigure}[c]{3.9cm}
        \includegraphics[width=4cm]{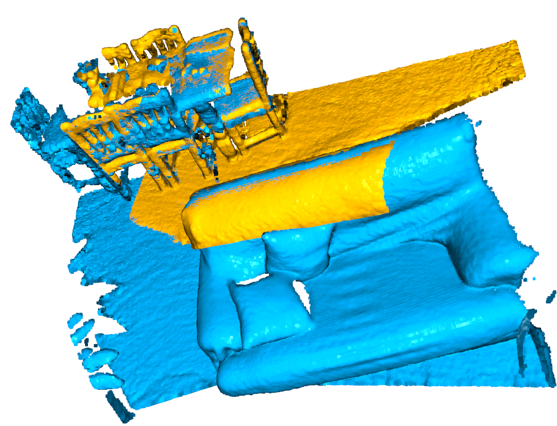}
        \subcaption{Home1}
    \end{subfigure}
    \\
    \begin{subfigure}[c]{3.9cm}
        \includegraphics[width=4.55cm]{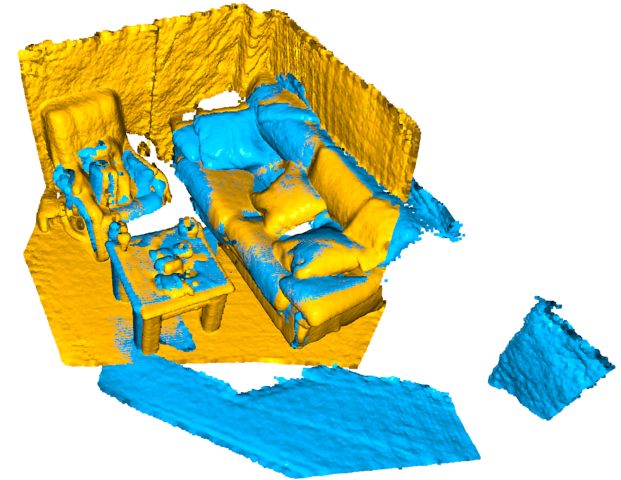}
        \subcaption{Home2}
    \end{subfigure}
    \quad
    \begin{subfigure}[c]{3.9cm}
        \includegraphics[width=4cm]{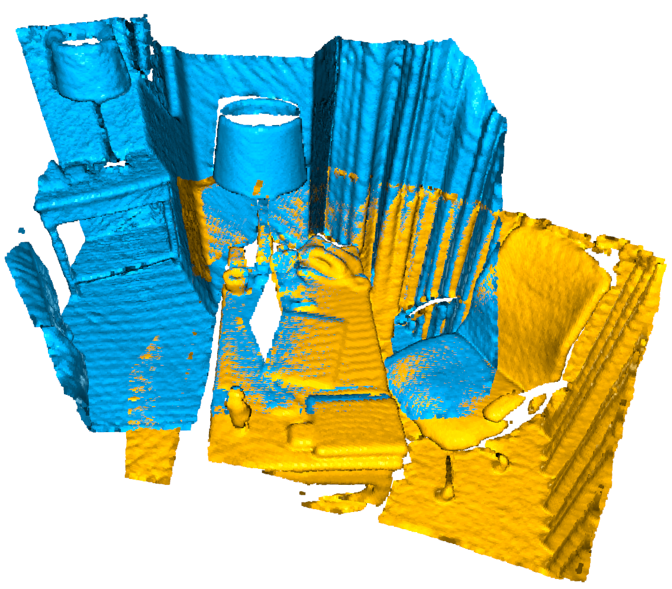}
        \subcaption{Hotel1}
    \end{subfigure}
    \\
    \begin{subfigure}[c]{3.9cm}
        \includegraphics[width=4.55cm]{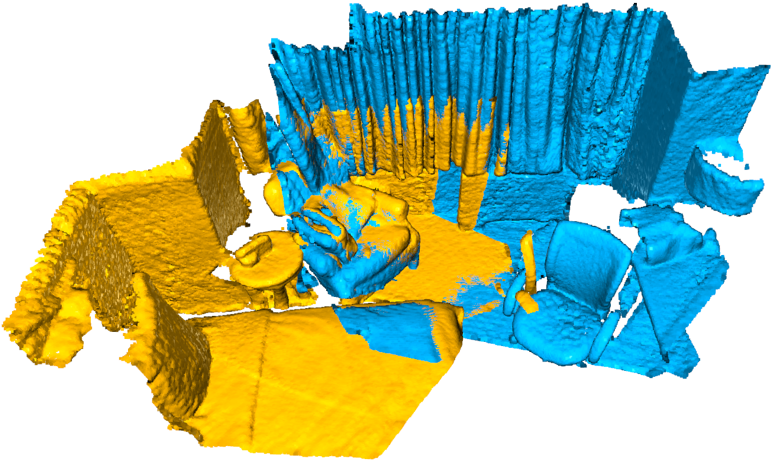}
        \subcaption{Hotel2}
    \end{subfigure}
    \quad
    \begin{subfigure}[c]{3.9cm}
        \includegraphics[width=4cm]{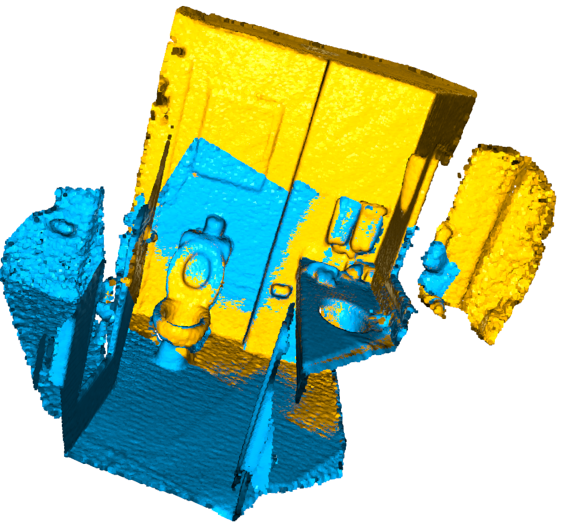}
        \subcaption{Hotel3}
    \end{subfigure}
    \\
    \begin{subfigure}[c]{3.9cm}
        \includegraphics[width=4.6cm]{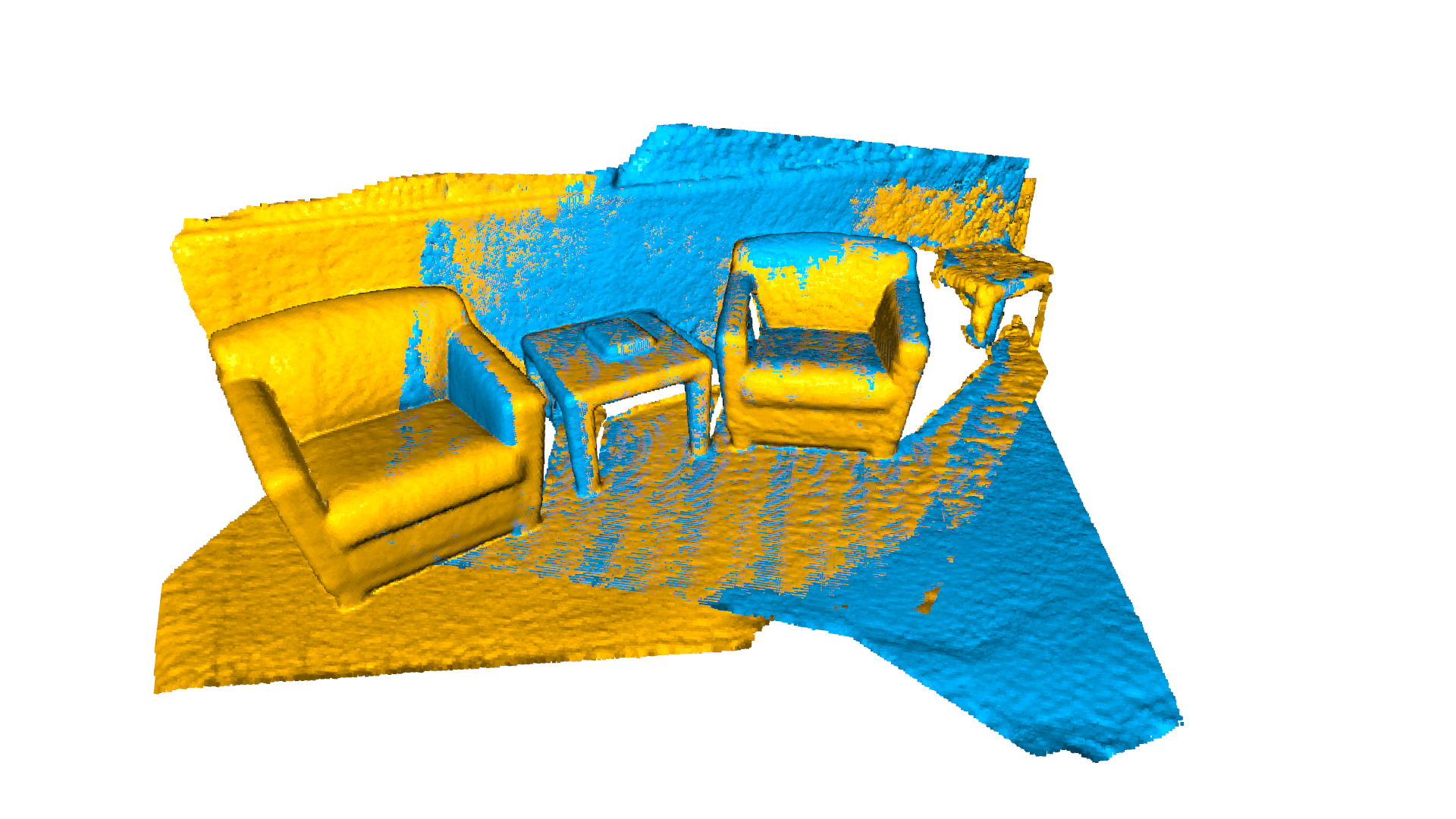}
        \subcaption{Study}
    \end{subfigure}
    \quad
    \begin{subfigure}[c]{3.9cm}
        \includegraphics[width=3.5cm]{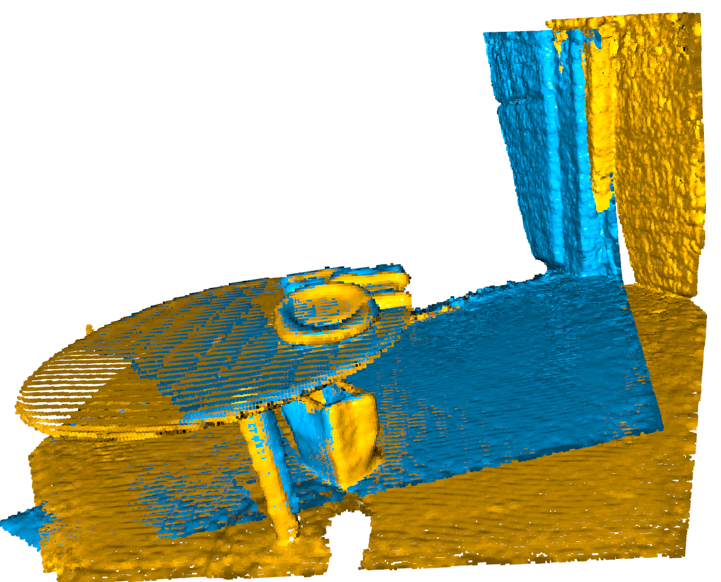}
        \subcaption{Lab}
    \end{subfigure}
    \caption{The registration visualization results of our registration method in 8 different scenes of the 3DMatch benchmark~(\citeauthor{3DMatch2017}, \citeyear{3DMatch2017}).}
    \label{FigVisulizationEachSceneResults}
\end{figure}

\begin{figure*}[hbt]
    \centering
    \includegraphics[width=17cm]{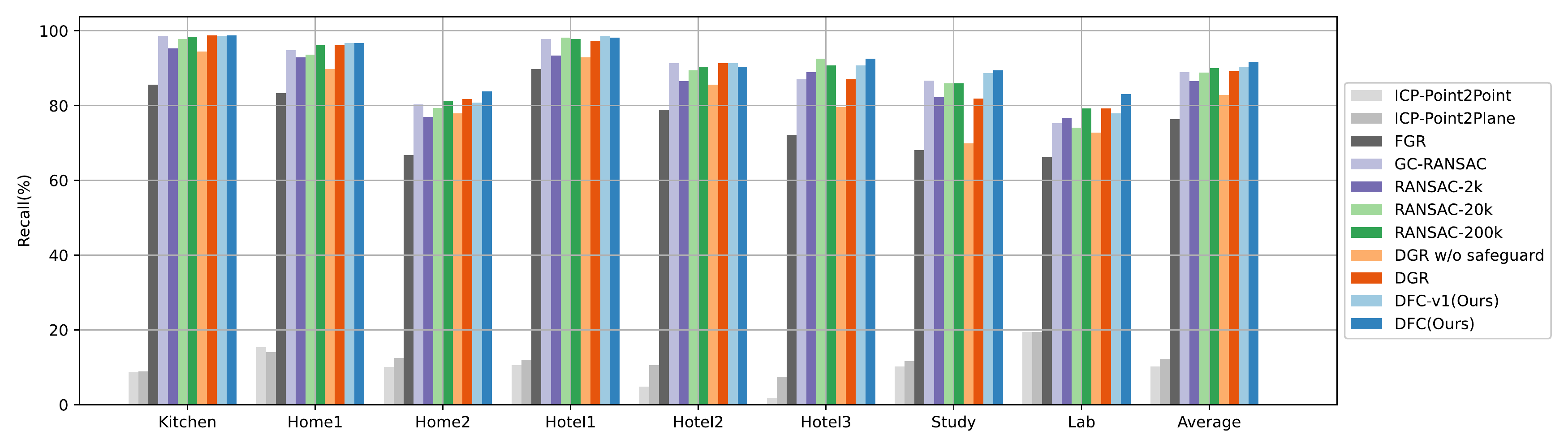}\\
    \includegraphics[width=17cm]{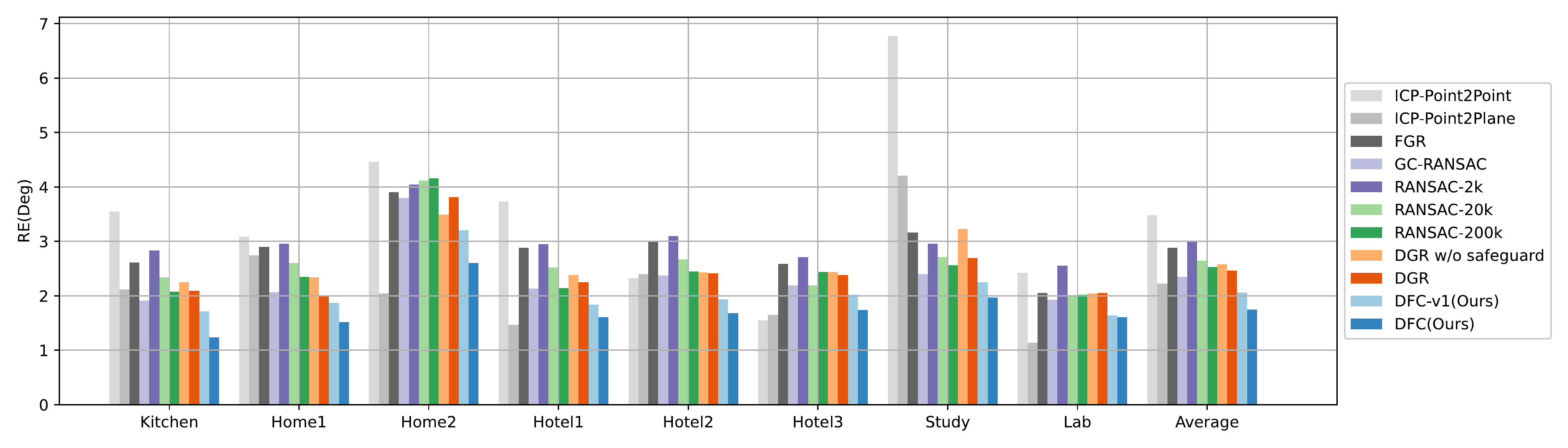}\\
    \includegraphics[width=17cm]{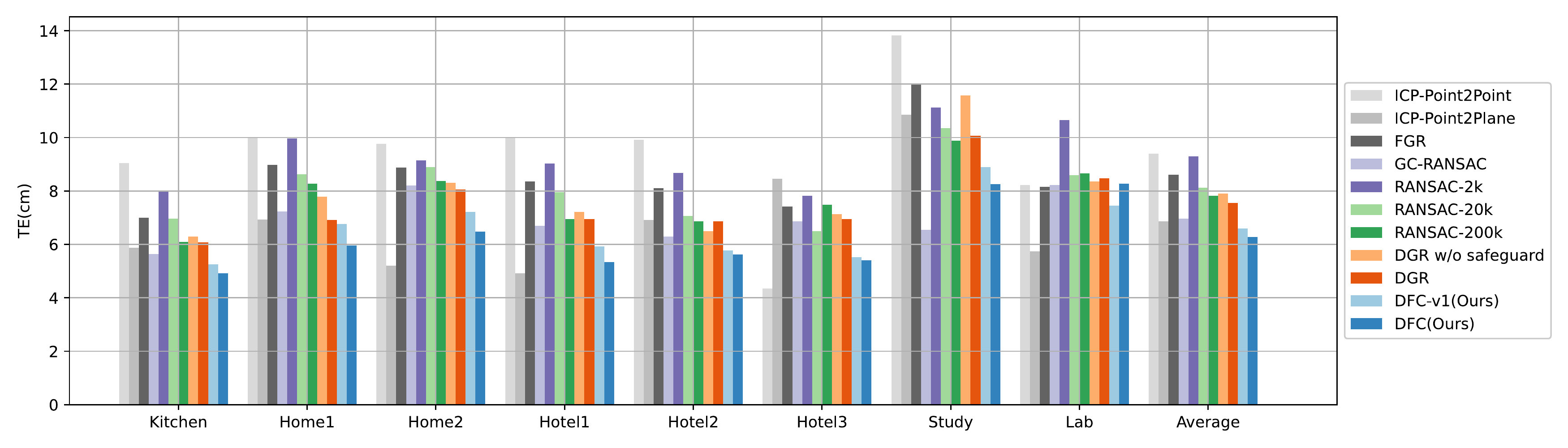}
    \caption{Registration results  per scene of the 3DMatch benchmark. \textit{Row 1-3}: \textit{Registration recall}~(higher is better), $\mathrm{TE}$ and $\mathrm{RE}$ measured on successfully aligned pairs (lower is better). Our method consistently performs  better on most scenes. The last column is the average \textit{registration recall}, $\mathrm{TE}$ and $\mathrm{RE}$ on all eight scenes. A certain missing bar means its value is zero and there are no successful alignments in the corresponding scene.}
    \label{FigRegistrationResultsonPerScene}
\end{figure*}

 
In this section, we compare and analyze the registration results on the test split of 3DMatch~(\citeauthor{3DMatch2017}, \citeyear{3DMatch2017}), which contains 8 different indoor scenes, as shown in Fig.~\ref{FigVisulizationEachSceneResults}. We further evaluate the performance of our method on the 3DMatch dataset by computing $\mathrm{RR}$, $\mathrm{RE}$ and $\mathrm{TE}$ based on Eq.~\ref{Eq_RE_TE}, respectively. After obtaining the initial rigid transformation, we attempt to use the ICP algorithm~(\citeauthor{ICP1992}, \citeyear{ICP1992}) for subsequent optimization of the initial predicted transformation during the testing stage. By a slight abuse of notation, we define our pipeline without subsequent ICP refinement as \textbf{DFC-v1} and the full model with ICP as \textbf{DFC}. Fig.~\ref{FigRegistrationResultsonPerScene} summarizes the detailed statistics on each test scene. Our method outperforms the other advanced classical and learning-based methods in terms of  recall ratio and reaches the lowest   $\mathrm{RE}$ and $\mathrm{TE}$ on most scenes, which implies that our method  has much more robustness when dealing with different real scenes. 
\textbf{Evaluation Metrics.} ~To make a fair comparison with other state-of-the-art registration methods~(\citeauthor{DeepClosetPoint2019}, \citeyear{DeepClosetPoint2019}; \citeauthor{DeepGlobalRegistration2020}, \citeyear{DeepGlobalRegistration2020}; \citeauthor{PointDSC2021}, \citeyear{PointDSC2021}; \citeauthor{FastGlobalRegistration2016}, \citeyear{FastGlobalRegistration2016}), we adopt the following three evaluation metrics to evaluate the performance of the proposed DFC method:
\par 
(1) Rotation error~($\mathrm{RE}$) and translation error~($\mathrm{TE}$). $\mathrm{RE}$ and $\mathrm{TE}$ penalize errors between estimated poses and ground-truth poses: 
\begin{align}
    \mathrm{RE}\left( \mathbf{R}^*,\mathbf{R}^{\mathbf{g}} \right) &=\mathrm{arc}\cos  \frac{\mathrm{Tr}\left( \left( \mathbf{R}^* \right) ^{-1}\mathbf{R}^{\mathbf{g}} \right) -1}{2}  
    \\
    \mathrm{TE}\left( \mathbf{t}^*,\mathbf{t}^{\mathbf{g}} \right) &=\left\| \mathbf{t}^*-\mathbf{t}^{\mathbf{g}} \right\| 
    \label{Eq_RE_TE}
\end{align}
where $\mathbf{R}^{\mathbf{g}}$ and $\mathbf{t}^{\mathbf{g}}$  represent the ground-truth rotation and translation, respectively, and $\mathrm{Tr}\left( \cdot \right)$  represents the trace of one certain matrix. It is worth noting that $\mathrm{RE}$ and $\mathrm{TE}$ are calculated only when two point clouds are successfully aligned. This is because two point clouds that fail to align will return an incorrect pose estimate that differs significantly from the ground-truth transformation and makes the predictions for $\mathrm{RE}$ and $\mathrm{TE}$ unreliable.
\par 
(2) Registration Recall~($\mathrm{RR}$)~(\citeauthor{RobustReconstructionofIndoorsScenes2015}, \citeyear{RobustReconstructionofIndoorsScenes2015}). The recall ratio metric represents the percentage of successful pairwise alignments. This means successful alignment when $\mathrm{TE}$ and $\mathrm{RE}$ are less than some thresholds at the same time. For the 3DMatch benchmark, the pairwise alignment result can be regarded as successful if $\mathrm{RE}<15^{\degree}$ and $\mathrm{TE}<30cm$.
\par
\textbf{Candidate inliers sampling.}~The second step of our pipeline is weighting and sampling a set of  correspondences referred to as candidate latent inliers in one shot, instead of randomly sampling minimal subsets iteratively, such as RANSAC~(\citeauthor{RANSAC1981}, \citeyear{RANSAC1981}). These selected candidate inliers are characterized by high probability, thus making them have higher probabilities of becoming inliers compared to other correspondences. 
\par 
Table~\ref{Table_PerformencewithDifferentLatentInliers} shows the comprehensive assessment results of the proposed DFC method with different numbers of samples. We conclude that the sampling strategy can greatly improve the \textit{registration recall}. To some extent, \textit{registration recall} is higher when the number of candidate inlier samples is smaller; specifically, $\mathrm{RR}$ is optimal when $N_{\mathcal{S}}$ is set to $200$, which indicates that taking a portion of the correspondences with high confidence as candidate inliers can greatly improve the alignment effect and filter out the outliers effectively. It may be possible to achieve even higher  \textit{registration recall}   by combining the \itshape{top-k} \upshape sampling operator scheme. We leave this task for future research.
\begin{table}[hbt]
    \centering
    \caption{The performance of our method with different candidate inlier sampling strategies.}
    \label{Table_PerformencewithDifferentLatentInliers}
    \begin{tabular}{p{1.5cm} p{23 pt}<{\centering} p{23 pt}<{\centering} p{23 pt}<{\centering} p{23 pt}<{\centering} p{23 pt}<{\centering} }
        \toprule
        \multirow{2}*{Metrics}&\multicolumn{5}{c}{$N_{\mathcal{S}}$}\\
        \cline{2-6}
        \multicolumn{1}{l}{}&100&200&300&400&500\\      
        \midrule 
        $\mathrm{RR(\%)}$&92.98&$\mathbf{93.47}$&93.41&93.22&93.10\\
        $\mathrm{RE(Deg)}$&1.69&$\mathbf{1.67}$&1.69&1.68&1.67\\
        $\mathrm{TE(cm)}$&6.07&$\mathbf{6.04}$&6.06&6.08&6.06\\
        \bottomrule 
    \end{tabular}
\end{table}
\par 
\begin{sloppypar}
\textbf{Traditional methods.}~To compare with traditional registration methods, we choose five typical traditional algorithms as benchmarks: FGR~(\citeauthor{FastGlobalRegistration2016}, \citeyear{FastGlobalRegistration2016}), RANSAC~(\citeauthor{RANSAC1981}, \citeyear{RANSAC1981}), GC-RANSAC~(\citeauthor{GCRANSAC2018}, \citeyear{GCRANSAC2018}), Point2Point-ICP and Point2Plane-ICP. All registration algorithms are  implemented with the Open3D library, except for GC-RANSAC.
The assessment results are shown in Table~\ref{Table_Reg_Resultson3DMatch}. 
\end{sloppypar}
\par 
ICP variants, including point-to-point and point-to-plane ICP methods, fail in most indoor scenes, because the ICP algorithm is highly dependent on the initial pose estimation so that it easily falls into the local optimum and partly because the overlap ratio between 3D scans is low. The lower the overlap between two point clouds is, the more likely the point clouds fail to align. The performances of FGR, RANSAC and GC-RANSAC are better than that of the ICP variants. The FGR network achieves a recall as high as $78.62\%$ when applied with FCGF feature descriptors, and even RANSAC reaches $91.99\%$ \textit{registration recall}. When aligning point clouds with 200k  sampling iterations, RANSAC can still maintain strong robustness. This conclusion is absolutely different from the literature~(\citeauthor{DeepGlobalRegistration2020}, \citeyear{DeepGlobalRegistration2020}), which is due to the  fast and compact characteristics of FCGF feature descriptors compared to classical FPFH descriptors. However, it is worth noting that our method is approximately 10 times faster than RANSAC-200k, and our method achieves much higher  \textit{registration recall} by a significant margin. Another method named GC-RANSAC only achieves a $91.68\%$ recall ratio, and our method exceeds $1.79\%$.
\par
\begin{table*}
    \centering
    \caption{Quantitative comparisons of different registration algorithms on the 3DMatch benchmark. Time excludes the construction of matched correspondences.}
    \label{Table_Reg_Resultson3DMatch}
    \begin{tabular}{p{7cm} p{2cm}<{\centering} p{2cm}<{\centering} p{2cm}<{\centering} p{2cm}<{\centering}}
        \toprule
        {Methods}&$\mathrm{RR(\%\uparrow)}$& $\mathrm{RE(deg\downarrow)}$&$\mathrm{TE(cm\downarrow)}$&$\mathrm{Time(s)}$\\
        \midrule
        FGR(\citeauthor{FastGlobalRegistration2016}, \citeyear{FastGlobalRegistration2016})&78.62&2.91&8.42&0.64\\
        RANSAC-2k(\citeauthor{RANSAC1981}, \citeyear{RANSAC1981})&88.42&3.02&9.14&0.15\\
        RANSAC-20k&91.13&2.67&8.03&1.05\\
        RANSAC-200k&91.99&2.47&7.53&10.89\\
        GC-RANSAC-1M(\citeauthor{GCRANSAC2018}, \citeyear{GCRANSAC2018}) &91.68&2.29&7.09&0.42\\
        \midrule
        ICP(Point2Point)(\citeauthor{Open3D2018}, \citeyear{Open3D2018})&10.10&4.06&10.21&0.10\\
        ICP(Point2Plane)(\citeauthor{Open3D2018}, \citeyear{Open3D2018})&11.34&2.40&6.79&0.71\\
        \midrule
        DGR w/o safeguard&85.20&2.58&7.73&0.70\\
        DGR(\citeauthor{DeepGlobalRegistration2020}, \citeyear{DeepGlobalRegistration2020})&91.30&2.43&7.34&1.21\\
        PointDSC(\citeauthor{PointDSC2021}, \citeyear{PointDSC2021})&93.28&2.06&6.55&0.09\\
        DFC-v1(Ours)&92.54&2.04&6.56&0.08\\
        DFC(Ours)&$\mathbf{93.47}$&$\mathbf{1.67}$&$\mathbf{6.04}$&0.14\\
        \bottomrule
    \end{tabular}
\end{table*}
\textbf{Learning-based methods.}
In addition, we choose another three state-of-the-art learning-based algorithms named 3DRegNet~(\citeauthor{3DRegNet2020}, \citeyear{3DRegNet2020}), DGR~(\citeauthor{DeepGlobalRegistration2020}, \citeyear{DeepGlobalRegistration2020}) and PointDSC~(\citeauthor{PointDSC2021}, \citeyear{PointDSC2021}) as our comparison benchmarks, and the registration results of DGR are also recorded in the absence of a protection mechanism (i.e., the RANSAC algorithm is used to optimize the initial poses during the evaluation phase). The  \textit{registration recall} of our method, especially without applying the ICP algorithm to optimize the estimated initial poses, has already exceeded $1.24\%$ compared to DGR. DGR only achieves a recall ratio of $86.2\%$ in the absence of RANSAC optimization~(called the safeguard mechanism). Compared with PointDSC, our method has a slightly higher \textit{registration recall} than PointDSC after optimizing the estimated poses using the ICP algorithm, and the model runtime is only $0.08s$ without ICP refinement. 
\par 
In conclusion, the DFC method proposed in this paper provides an efficient and robust registration method in terms of \textit{registration recall}, and achieves a better balance between efficient computation and robustness at the same time. 

\subsection{Multiway Registration}
\begin{table*}[hbt]
    \centering
    \caption{ATE(cm) on the augmented ICL-NUIM dataset with simulated depth noise. The last column shows the average ATE of all four scenes. For BAD-SLAM, this method fails in the scene "Living room 1", so we do not compute its average ATE.}
    \label{Table_MultiwayRegistration}
    \begin{tabular}{p{7cm} p{1.6cm}<{\centering} p{1.6cm}<{\centering} p{1.6cm}<{\centering} p{1.6cm}<{\centering} p{1.6cm}<{\centering}}
        \toprule
        {Method}&Living1&Living2&Office1&Office2&Average\\
        \midrule
        ElasticFusion(\citeauthor{ElasticFusion2015}, \citeyear{ElasticFusion2015}) &66.61&24.33&13.04&35.02&34.75\\
        InfiniTAM(\citeauthor{InfiniTAM2016}, \citeyear{InfiniTAM2016})&46.07&73.64&113.8&105.2&84.68\\
        BAD-SLAM(\citeauthor{BadSlam2019}, \citeyear{BadSlam2019})&-&40.41&18.53&26.34&-\\
        Multiway+FGR(\citeauthor{FastGlobalRegistration2016}, \citeyear{FastGlobalRegistration2016})&78.97&24.91&14.96&21.05&34.98\\
        Multiway+RANSAC(\citeauthor{RANSAC1981}, \citeyear{RANSAC1981})&110.9&19.33&14.42&17.31&40.49\\
        Multiway+DGR(\citeauthor{DeepGlobalRegistration2020}, \citeyear{DeepGlobalRegistration2020})&21.06&21.88&15.76&11.56&17.57\\
        Multiway+PointDSC(\citeauthor{PointDSC2021}, \citeyear{PointDSC2021})&20.25&15.58&13.56&$\mathbf{11.30}$&$\mathbf{15.18}$\\
        Multiway+DFC(Ours)&$\mathbf{18.15}$&$\mathbf{15.28}$&$\mathbf{12.76}$&32.44&19.66\\
        \bottomrule
    \end{tabular}
\end{table*}
\par  
To evaluate the generalization capability of our method to new datasets, we use the training model on the 3DMatch dataset again and then evaluate the performance of multiway registration on the augmented ICL-NUIM dataset, which is also called cross-dataset generalization capability analysis. 
\par 
Following~(\citeauthor{PointDSC2021}, \citeyear{PointDSC2021}; \citeauthor{DeepGlobalRegistration2020}, \citeyear{DeepGlobalRegistration2020}), we apply our method to roughly align all scan fragments and to obtain the initial poses and then trim the initial poses with multiway registration using the pose graph optimization algorithm which can be implemented with the open-source library Open3D~(\citeauthor{Open3D2018}, \citeyear{Open3D2018}). To assess multiway registration, we measure the absolute trajectory error~(ATE) on the augmented ICL-NUIM dataset with simulated depth noise, and the results are shown in Table~\ref{Table_MultiwayRegistration}. Compared  with current state-of-the-art online SLAM methods and offline reconstruction systems, our method achieves the lowest level of ATE in the first three scenes. 
\subsection{Outdoor LIDAR Registration}

\begin{figure}[hbt]
    \centering
    \begin{subfigure}{7cm}
        \includegraphics[scale=0.1]{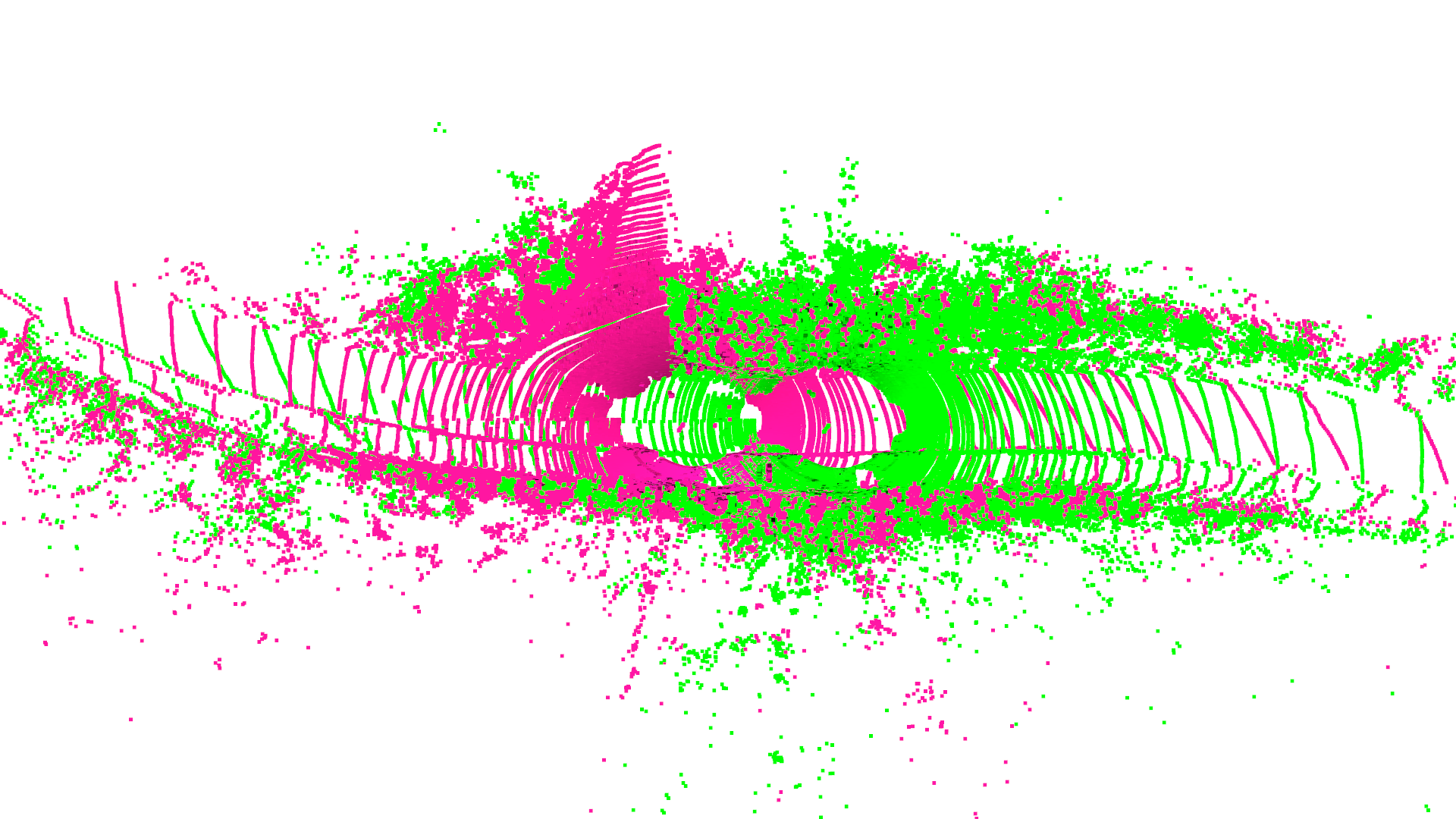}
        \subcaption{KITTI registration test pair 1}
    \end{subfigure}
    \begin{subfigure}{7cm}
        \includegraphics[scale=0.1]{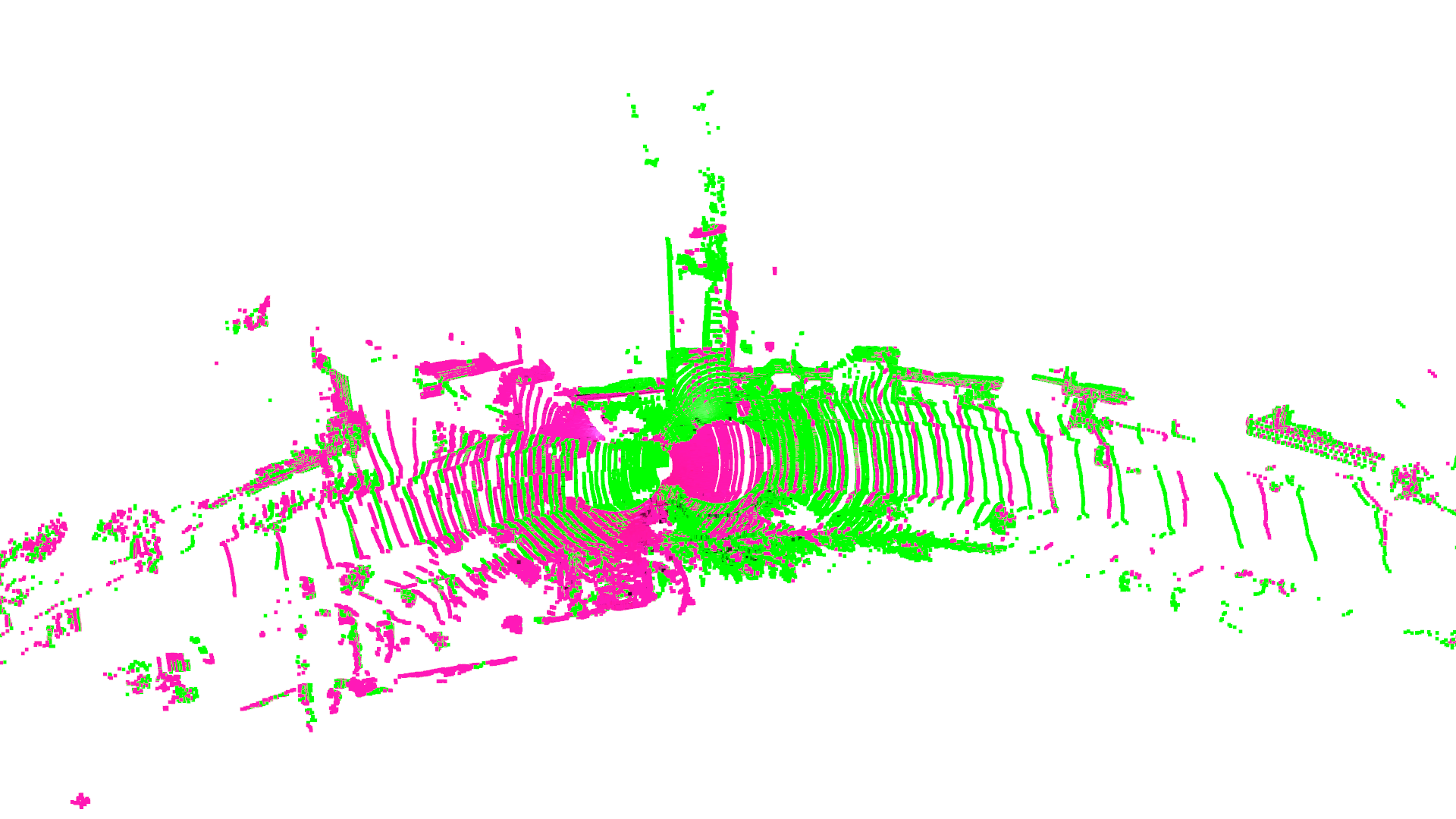}
        \subcaption{KITTI registration test pair 2}
    \end{subfigure}
    \begin{subfigure}{7cm}
        \includegraphics[scale=0.1]{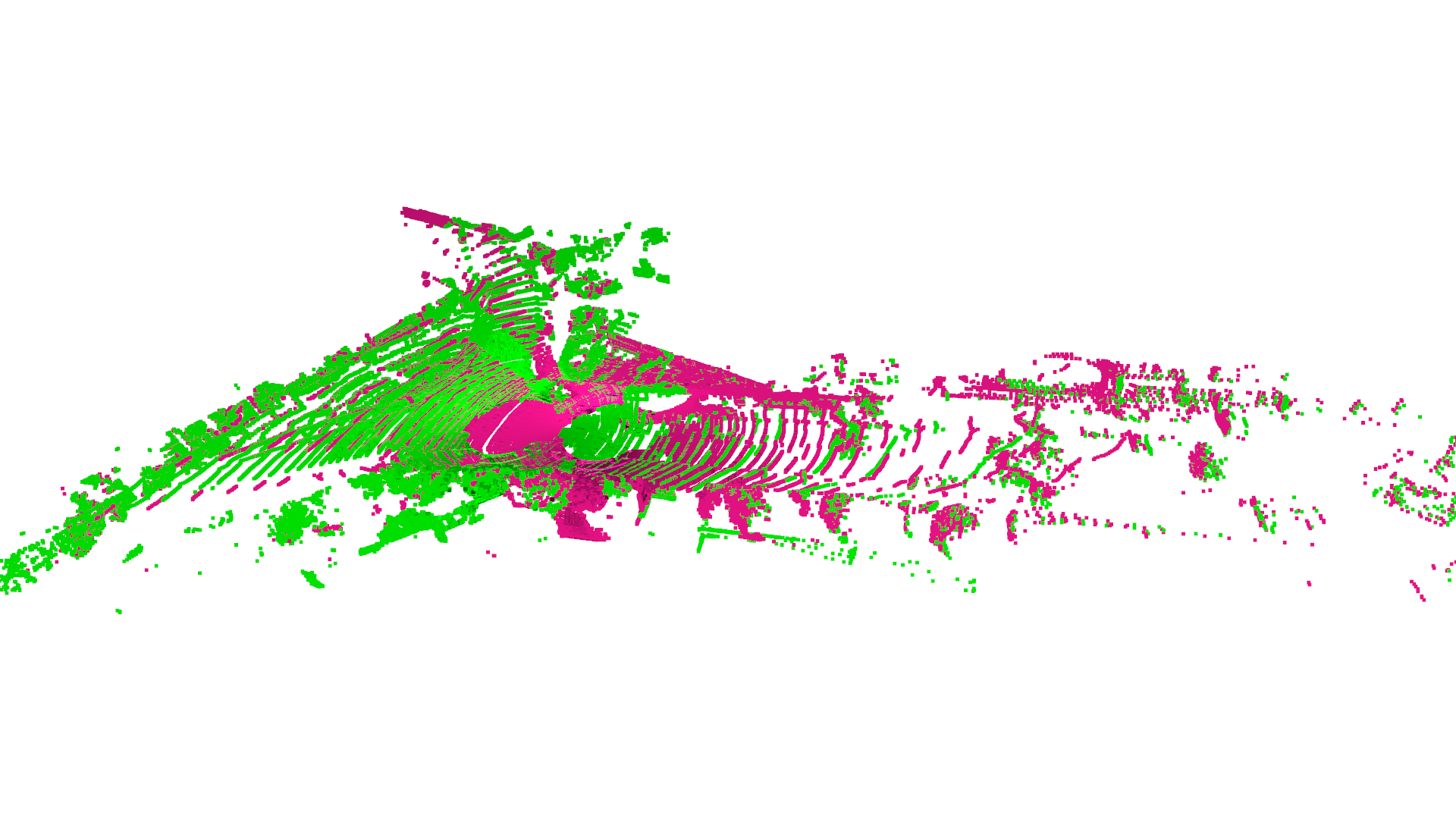}
        \subcaption{KITTI registration test pair 3}
    \end{subfigure}
    \begin{subfigure}{7cm}
        \includegraphics[scale=0.1]{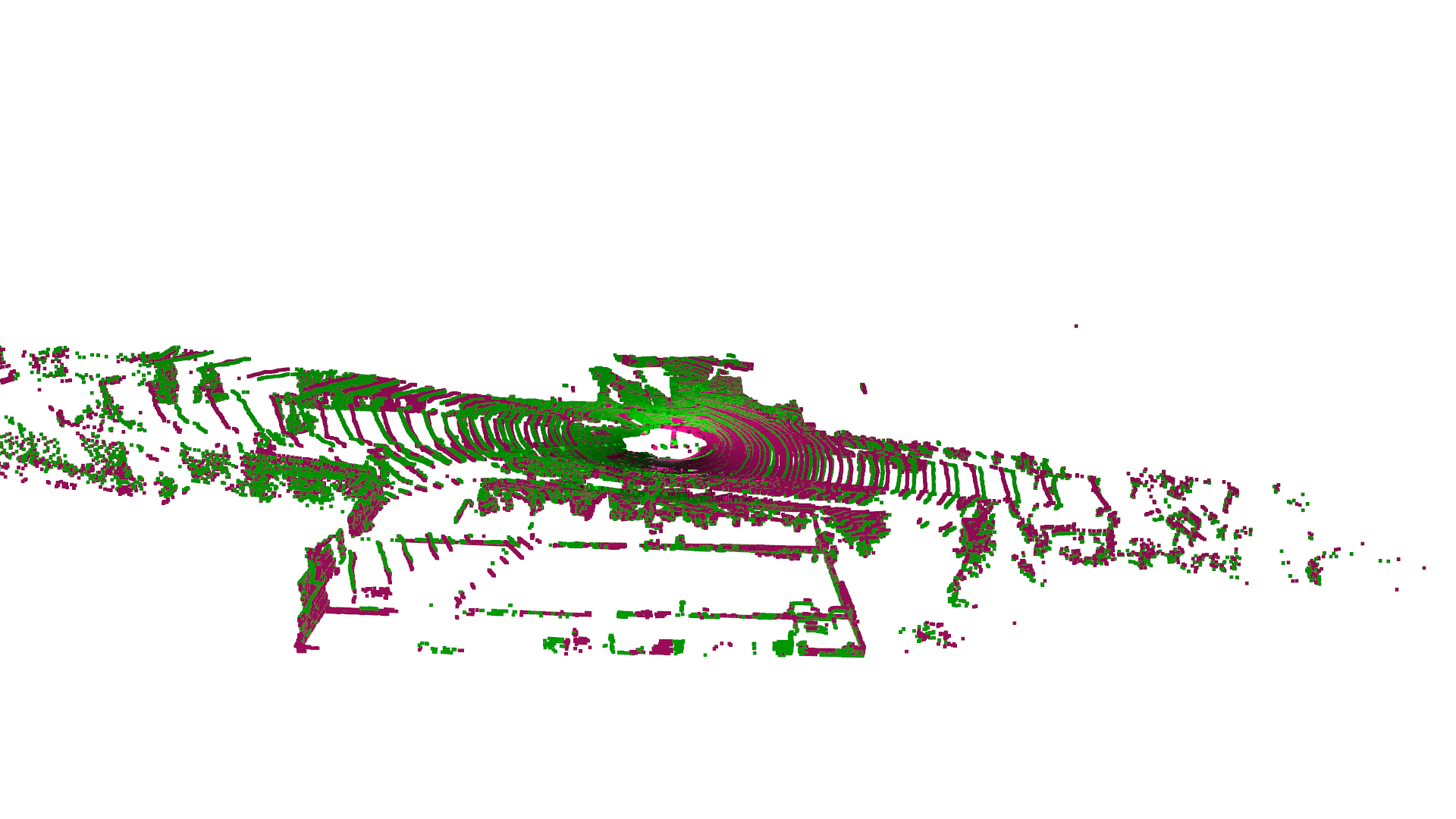}
        \subcaption{KITTI registration test pair 4}
    \end{subfigure}
    \caption{Registration visulization examples on KITTI. Our method can effectively and accurately align the outdoor point cloud scans.}
    \label{Fig_VilulizationRegResultsonKITTI}
\end{figure}

\begin{table*}
    \centering
    \caption{Quantitative comparisons of different registration algorithms on KITTI odometry. The listed time excludes the construction of the correspondence set.}
    \label{Table_Reg_ResultsonKitti}
    \begin{tabular}{p{7cm} p{2cm}<{\centering} p{2cm}<{\centering} p{2cm}<{\centering} p{2cm}<{\centering}}
        \toprule
        {Methods}&$\mathrm{RR(\%\uparrow)}$& $\mathrm{RE(deg\downarrow)}$&$\mathrm{TE(cm\downarrow)}$&$\mathrm{Time(s)}$\\
        \midrule
        RANSAC-1k(\citeauthor{RANSAC1981}, \citeyear{RANSAC1981})&96.58&0.48&23.41&0.30\\
        RANSAC-20k&97.48&0.38&22.60&3.59\\
        RANSAC-200k&97.12&0.35&22.32&37.44\\
        GCRANSAC-2k(\citeauthor{GCRANSAC2018}, \citeyear{GCRANSAC2018})&96.22&0.44&23.33&0.51\\
        \midrule
        DGR(\citeauthor{DeepGlobalRegistration2020}, \citeyear{DeepGlobalRegistration2020})&96.90&0.33&21.29&0.86\\
        PointDSC(\citeauthor{PointDSC2021}, \citeyear{PointDSC2021})&$\mathbf{98.20}$&0.33&20.94&0.31\\
        DFC(Ours)&$97.30$&$\mathbf{0.24}$&$\mathbf{18.64}$&0.55\\
        \bottomrule
    \end{tabular}
\end{table*}
Following~(\citeauthor{FullyConvolutionalGeometricFeatures2019}, \citeyear{FullyConvolutionalGeometricFeatures2019}), we use outdoor LiDAR scans from the KITTI odometry~(\citeauthor{KITTIDataset2012}, \citeyear{KITTIDataset2012}) for registration. Similar to pairwise registration in indoor scenes, we set the inlier threshold $\tau$ to $60cm$, downsample the point clouds  using a voxelization filter with a $30cm$ voxel and then extract the  pointwise features by applying FCGF descriptors to form the correspondence set as the input of our pipeline. As illustrated in the literature~(\citeauthor{DeepGlobalRegistration2020}, \citeyear{DeepGlobalRegistration2020}; \citeauthor{PointDSC2021}, \citeyear{PointDSC2021}; \citeauthor{FullyConvolutionalGeometricFeatures2019}, \citeyear{FullyConvolutionalGeometricFeatures2019}), the thresholds of $\mathrm{TE}$ and $\mathrm{RE}$ are set to $60cm$ and  $5^{\degree}$, respectively. When both $\mathrm{TE}$ and $\mathrm{RE}$ are less than the abovementioned thresholds, the alignment of two point clouds in KITTI odometry can be considered successful. The quantified results of our network on the KITTI odometry dataset are shown in Table~\ref{Table_Reg_ResultsonKitti}, and the visualization results are shown in Fig.~\ref{Fig_VilulizationRegResultsonKITTI}. Although the recall ratio of our method is slightly below another learning-based method PointDSC, our network still appears to be strongly competitive in decreasing the transformation errors.
\subsection{Ablation Studies}\label{sec5AblationStudies}
We designed some ablation experiments by dividing our pipeline into multiple modules and replacing each module with another different counterpart to study the role of each part. All experiments are performed on the same condition as the experiments in Sec~\ref{Sec5.2PairwiseRegistration}. 
\par  
\textbf{Ablation on feature embedding.}
To discuss the effectiveness of the proposed multiscale GFM network, we design ablation experiments on the 3DMatch benchmark to make a comparison. We choose two methods: PointNet and DGCNN. The comparison results are shown in Table~\ref{TableAblationstudyofMulti-ScaleGFM}. PointNet~(\citeauthor{PointNet2017}, \citeyear{PointNet2017}) achieves state-of-the-art performance for classification and segmentation and provides a new research perspective on processing raw point clouds while DGCNN~(\citeauthor{DGCNN2019}, \citeyear{DGCNN2019}) constructs  local  graph geometric features by using a local neighborhood graph and convolution operations on edges. Our method performs consistently better with the  proposed multiscale GFM network than PointNet and DGCNN, which provides strong evidence that our multiscale GFM module can help to improve the recall ratio.
\begin{table}[H]
    \centering
    \caption{Ablation study on the multiscale GFM network.}
    \label{TableAblationstudyofMulti-ScaleGFM}
    \begin{tabular}{p{0.8cm}  p{1.2cm}<{\centering} p{1.2cm}<{\centering} p{1.2cm}<{\centering} p{1.2cm}<{\centering} p{0.6cm} }
        \toprule
        \makecell[c]{Metrics}& \makecell[c]{PointNet+\\DFC-v1}& \makecell[c]{PointNet+\\DFC}& \makecell[c]{DGCNN+\\DFC-v1}& \makecell[c]{DGCNN+\\DFC}& \makecell[c]{Ours} \\
        \midrule
        $\mathrm{RR}$&91.13&91.68&92.73&93.22&$\mathbf{93.47}$\\
        $\mathrm{RE}$&2.03&1.67&2.05&1.68&$\mathbf{1.67}$ \\
        $\mathrm{TE}$&6.41&$\mathbf{6.01}$&6.50&6.02&6.04\\
        \bottomrule
    \end{tabular}
\end{table}
\par 
\textbf{Ablation on principal vectors.} In this experiment, we subsequently design ablation experiments to explore whether the eigenvalue or PCA method is better for registration. We compare PCA and Eigenvalue with both DFC-v1 and DFC networks on the 3DMatch benchmark. Table~\ref{TableAblationstudyofPrincipalVectors} shows both DFC-v1 and DFC perform better with PCA than Eigenvalue. Therefore, we can conclude that the PCA algorithm is helpful to boost the registration performance.
\begin{table}[hbt]
    \centering
    \caption{Ablation study on  principal vectors.}
    \label{TableAblationstudyofPrincipalVectors}
    \begin{tabular}{p{0.8cm}  p{1.5cm}<{\centering} p{1.5cm}<{\centering} p{0.9cm}<{\centering} p{0.9cm}<{\centering}}
        \toprule
        \makecell[c]{Metrics}& \makecell[c]{Eigenvalue+\\DFC-v1}& \makecell[c]{Eigenvalue+\\DFC}& \makecell[c]{PCA+\\DFC-v1}& \makecell[c]{PCA+\\DFC}\\
        \midrule
        $\mathrm{RR}$&92.17&92.79&92.73&$\mathbf{93.47}$\\
        $\mathrm{RE}$&2.02&$\mathbf{1.65}$&2.05&1.67 \\
        $\mathrm{TE}$&6.54&$\mathbf{6.02}$&6.50&6.04\\
        \bottomrule
    \end{tabular}
\end{table}

%% file: includefiles/section06.tex
\section{Conclusion}\label{sec06Conclusion}
We propose deep feature consistency, a learning-based framework for robust, accurate and efficient point cloud registration by jointly embedding the correspondence features with a multiscale GFM network, weighting and sampling correspondences with a correspondence weighting module, and solving the rigid transformation for alignment with a deep feature matching module. 
The results on 3DMatch indoor scenes and KITTI odometry outdoor scenes show that our methodology outperforms the traditional and learning-based algorithms and can effectively and quickly align the point clouds. The low transformation errors and high robustness of our method make it attractive for many applications relying on the point cloud registration task. In a further extension of this work, we will explore new methods for improving the generalization capability in broader application scenarios and attempt to extend our method to other 3D computer visual tasks, such as 3D reconstruction and mapping, object pose estimation.

%% file: DeepFeatureConsistency.bbl
\begin{thebibliography}{56}
\expandafter\ifx\csname natexlab\endcsname\relax\def\natexlab#1{#1}\fi
\providecommand{\url}[1]{\texttt{#1}}
\providecommand{\href}[2]{#2}
\providecommand{\path}[1]{#1}
\providecommand{\DOIprefix}{doi:}
\providecommand{\ArXivprefix}{arXiv:}
\providecommand{\URLprefix}{URL: }
\providecommand{\Pubmedprefix}{pmid:}
\providecommand{\doi}[1]{\href{http://dx.doi.org/#1}{\path{#1}}}
\providecommand{\Pubmed}[1]{\href{pmid:#1}{\path{#1}}}
\providecommand{\bibinfo}[2]{#2}
\ifx\xfnm\relax \def\xfnm[#1]{\unskip,\space#1}\fi
\bibitem[{Ahmed et~al.(2019)Ahmed, Das and
  Chaudhury}]{Least_squares_registration2019}
\bibinfo{author}{Ahmed, S.M.}, \bibinfo{author}{Das, N.R.},
  \bibinfo{author}{Chaudhury, K.N.}, \bibinfo{year}{2019}.
\newblock \bibinfo{title}{Least-squares registration of point sets over se(d)
  using closed-form projections}.
\newblock \bibinfo{journal}{Computer Vision and Image Understanding}
  \bibinfo{volume}{183}, \bibinfo{pages}{20--32}.
\newblock \DOIprefix\doi{10.1016/j.cviu.2019.03.008}.
\bibitem[{An et~al.(2022)An, Liang, Yu, Fang and
  Ma}]{Deep_structural_Information_fusion2022}
\bibinfo{author}{An, P.}, \bibinfo{author}{Liang, J.}, \bibinfo{author}{Yu,
  K.}, \bibinfo{author}{Fang, B.}, \bibinfo{author}{Ma, J.},
  \bibinfo{year}{2022}.
\newblock \bibinfo{title}{Deep structural information fusion for 3d object
  detection on lidar–camera system}.
\newblock \bibinfo{journal}{Computer Vision and Image Understanding}
  \bibinfo{volume}{214}, \bibinfo{pages}{103295}.
\newblock \DOIprefix\doi{10.1016/j.cviu.2021.103295}.
\bibitem[{Aoki et~al.(2019)Aoki, Goforth, Srivatsan and Lucey}]{PointNetLK2019}
\bibinfo{author}{Aoki, Y.}, \bibinfo{author}{Goforth, H.},
  \bibinfo{author}{Srivatsan, R.A.}, \bibinfo{author}{Lucey, S.},
  \bibinfo{year}{2019}.
\newblock \bibinfo{title}{Pointnetlk: Robust \& efficient point cloud
  registration using pointnet}.
\bibitem[{Bai et~al.(2021)Bai, Luo, Zhou, Chen, Li, Hu, Fu and
  Tai}]{PointDSC2021}
\bibinfo{author}{Bai, X.}, \bibinfo{author}{Luo, Z.}, \bibinfo{author}{Zhou,
  L.}, \bibinfo{author}{Chen, H.}, \bibinfo{author}{Li, L.},
  \bibinfo{author}{Hu, Z.}, \bibinfo{author}{Fu, H.}, \bibinfo{author}{Tai,
  C.L.}, \bibinfo{year}{2021}.
\newblock \bibinfo{title}{Pointdsc: Robust point cloud registration using deep
  spatial consistency}, in: \bibinfo{booktitle}{Proceedings of the IEEE/CVF
  Conference on Computer Vision and Pattern Recognition}, pp.
  \bibinfo{pages}{15859--15869}.
\bibitem[{Barath and Matas(2018)}]{GCRANSAC2018}
\bibinfo{author}{Barath, D.}, \bibinfo{author}{Matas, J.},
  \bibinfo{year}{2018}.
\newblock \bibinfo{title}{Graph-cut ransac}, in:
  \bibinfo{booktitle}{Proceedings of the IEEE conference on computer vision and
  pattern recognition}, pp. \bibinfo{pages}{6733--6741}.
\bibitem[{Besl and McKay(1992)}]{ICP1992}
\bibinfo{author}{Besl, P.J.}, \bibinfo{author}{McKay, N.D.},
  \bibinfo{year}{1992}.
\newblock \bibinfo{title}{A method for registration of 3-d shapes}, in:
  \bibinfo{booktitle}{Sensor fusion IV: control paradigms and data structures},
  \bibinfo{publisher}{International Society for Optics and Photonics}. pp.
  \bibinfo{pages}{586--606}.
\bibitem[{Bustos et~al.(2019)Bustos, Chin, Neumann, Friedrich and
  Katzmann}]{maximumclique2019}
\bibinfo{author}{Bustos, A.P.}, \bibinfo{author}{Chin, T.J.},
  \bibinfo{author}{Neumann, F.}, \bibinfo{author}{Friedrich, T.},
  \bibinfo{author}{Katzmann, M.}, \bibinfo{year}{2019}.
\newblock \bibinfo{title}{A practical maximum clique algorithm for matching
  with pairwise constraints}.
\newblock \bibinfo{journal}{arXiv preprint arXiv:1902.01534}
  \bibinfo{volume}{2}.
\bibitem[{Cavalli et~al.(2020)Cavalli, Larsson, Oswald, Sattler and
  Pollefeys}]{Adalam2020}
\bibinfo{author}{Cavalli, L.}, \bibinfo{author}{Larsson, V.},
  \bibinfo{author}{Oswald, M.R.}, \bibinfo{author}{Sattler, T.},
  \bibinfo{author}{Pollefeys, M.}, \bibinfo{year}{2020}.
\newblock \bibinfo{title}{Adalam: Revisiting handcrafted outlier detection}.
\newblock \bibinfo{journal}{arXiv preprint arXiv:2006.04250} .
\bibitem[{Chang et~al.(2020)Chang, Ahn, Lee and
  Oh}]{Graph_matching_based_correspondence_search}
\bibinfo{author}{Chang, S.}, \bibinfo{author}{Ahn, C.}, \bibinfo{author}{Lee,
  M.}, \bibinfo{author}{Oh, S.}, \bibinfo{year}{2020}.
\newblock \bibinfo{title}{Graph-matching-based correspondence search for
  nonrigid point cloud registration}.
\newblock \bibinfo{journal}{Computer Vision and Image Understanding}
  \bibinfo{volume}{192}, \bibinfo{pages}{102899}.
\newblock \DOIprefix\doi{10.1016/j.cviu.2019.102899}.
\bibitem[{Charles et~al.(2017)Charles, Su, Kaichun and Guibas}]{PointNet2017}
\bibinfo{author}{Charles, R.Q.}, \bibinfo{author}{Su, H.},
  \bibinfo{author}{Kaichun, M.}, \bibinfo{author}{Guibas, L.J.},
  \bibinfo{year}{2017}.
\newblock \bibinfo{title}{Pointnet: Deep learning on point sets for 3d
  classification and segmentation}.
\newblock \DOIprefix\doi{10.1109/CVPR.2017.16}.
\bibitem[{Choi et~al.(2015)Choi, Zhou and
  Koltun}]{RobustReconstructionofIndoorsScenes2015}
\bibinfo{author}{Choi, S.}, \bibinfo{author}{Zhou, Q.Y.},
  \bibinfo{author}{Koltun, V.}, \bibinfo{year}{2015}.
\newblock \bibinfo{title}{Robust reconstruction of indoor scenes}, in:
  \bibinfo{booktitle}{Proceedings of the IEEE Conference on Computer Vision and
  Pattern Recognition}, pp. \bibinfo{pages}{5556--5565}.
\bibitem[{Choy et~al.(2020)Choy, Dong and Koltun}]{DeepGlobalRegistration2020}
\bibinfo{author}{Choy, C.}, \bibinfo{author}{Dong, W.},
  \bibinfo{author}{Koltun, V.}, \bibinfo{year}{2020}.
\newblock \bibinfo{title}{Deep global registration}, in:
  \bibinfo{booktitle}{2020 IEEE/CVF Conference on Computer Vision and Pattern
  Recognition (CVPR)}, \bibinfo{publisher}{IEEE},
  \bibinfo{address}{Piscataway,NJ}. pp. \bibinfo{pages}{2511--2520}.
\bibitem[{Choy et~al.(2019)Choy, Park and
  Koltun}]{FullyConvolutionalGeometricFeatures2019}
\bibinfo{author}{Choy, C.}, \bibinfo{author}{Park, J.},
  \bibinfo{author}{Koltun, V.}, \bibinfo{year}{2019}.
\newblock \bibinfo{title}{Fully convolutional geometric features}, in:
  \bibinfo{booktitle}{2019 IEEE/CVF International Conference on Computer Vision
  (ICCV)}, \bibinfo{publisher}{IEEE}, \bibinfo{address}{Piscataway,NJ}.
\bibitem[{Feng et~al.(2018)Feng, Zhang, Zhao, Ji and Gao}]{GVCNN2018}
\bibinfo{author}{Feng, Y.}, \bibinfo{author}{Zhang, Z.}, \bibinfo{author}{Zhao,
  X.}, \bibinfo{author}{Ji, R.}, \bibinfo{author}{Gao, Y.},
  \bibinfo{year}{2018}.
\newblock \bibinfo{title}{Gvcnn: Group-view convolutional neural networks for
  3d shape recognition}, in: \bibinfo{booktitle}{Proceedings of the IEEE
  Conference on Computer Vision and Pattern Recognition}, pp.
  \bibinfo{pages}{264--272}.
\bibitem[{Fischler and Bolles(1981)}]{RANSAC1981}
\bibinfo{author}{Fischler, M.A.}, \bibinfo{author}{Bolles, R.C.},
  \bibinfo{year}{1981}.
\newblock \bibinfo{title}{Random sample consensus}.
\newblock \bibinfo{journal}{Communications of the ACM} \bibinfo{volume}{24},
  \bibinfo{pages}{381--395}.
\bibitem[{Geiger et~al.(2012)Geiger, Lenz and Urtasun}]{KITTIDataset2012}
\bibinfo{author}{Geiger, A.}, \bibinfo{author}{Lenz, P.},
  \bibinfo{author}{Urtasun, R.}, \bibinfo{year}{2012}.
\newblock \bibinfo{title}{Are we ready for autonomous driving? the kitti vision
  benchmark suite}, in: \bibinfo{booktitle}{2012 IEEE Conference on Computer
  Vision and Pattern Recognition}, \bibinfo{publisher}{IEEE}. pp.
  \bibinfo{pages}{3354--3361}.
\bibitem[{Handa et~al.(2014)Handa, Whelan, Mcdonald and Davison}]{ICL-NUIM2014}
\bibinfo{author}{Handa, A.}, \bibinfo{author}{Whelan, T.},
  \bibinfo{author}{Mcdonald, J.}, \bibinfo{author}{Davison, A.J.},
  \bibinfo{year}{2014}.
\newblock \bibinfo{title}{A benchmark for rgb-d visual odometry, 3d
  reconstruction and slam}, in: \bibinfo{booktitle}{2014 IEEE International
  Conference on Robotics and Automation (ICRA)}, \bibinfo{publisher}{IEEE}.
\bibitem[{Hou et~al.(2017)Hou, Cheng, Hu, Borji, Tu and
  Torr}]{DeeplySupervisedSalientObjectDetection2017}
\bibinfo{author}{Hou, Q.}, \bibinfo{author}{Cheng, M.M.}, \bibinfo{author}{Hu,
  X.}, \bibinfo{author}{Borji, A.}, \bibinfo{author}{Tu, Z.},
  \bibinfo{author}{Torr, P.H.}, \bibinfo{year}{2017}.
\newblock \bibinfo{title}{Deeply supervised salient object detection with short
  connections}, in: \bibinfo{booktitle}{Proceedings of the IEEE conference on
  computer vision and pattern recognition}, pp. \bibinfo{pages}{3203--3212}.
\bibitem[{Huang et~al.(2021)Huang, Gojcic, Usvyatsov, Wieser and
  Schindler}]{PREDATOR2021}
\bibinfo{author}{Huang, S.}, \bibinfo{author}{Gojcic, Z.},
  \bibinfo{author}{Usvyatsov, M.}, \bibinfo{author}{Wieser, A.},
  \bibinfo{author}{Schindler, K.}, \bibinfo{year}{2021}.
\newblock \bibinfo{title}{Predator: Registration of 3d point clouds with low
  overlap}, in: \bibinfo{booktitle}{Proceedings of the IEEE/CVF Conference on
  Computer Vision and Pattern Recognition}, pp. \bibinfo{pages}{4267--4276}.
\bibitem[{Ioffe and Szegedy(2015)}]{BatchNormalization2015}
\bibinfo{author}{Ioffe, S.}, \bibinfo{author}{Szegedy, C.},
  \bibinfo{year}{2015}.
\newblock \bibinfo{title}{Batch normalization: Accelerating deep network
  training by reducing internal covariate shift}.
\bibitem[{Izatt et~al.(2020)Izatt, Dai and
  Tedrake}]{MixedIntegerProgramming2020}
\bibinfo{author}{Izatt, G.}, \bibinfo{author}{Dai, H.},
  \bibinfo{author}{Tedrake, R.}, \bibinfo{year}{2020}.
\newblock \bibinfo{title}{Globally Optimal Object Pose Estimation in Point
  Clouds with Mixed-Integer Programming}. \bibinfo{publisher}{Springer
  International Publishing}.
\newblock pp. \bibinfo{pages}{695--710}.
\bibitem[{Kähler et~al.(2016)Kähler, Prisacariu and Murray}]{InfiniTAM2016}
\bibinfo{author}{Kähler, O.}, \bibinfo{author}{Prisacariu, V.A.},
  \bibinfo{author}{Murray, D.W.}, \bibinfo{year}{2016}.
\newblock \bibinfo{title}{Real-Time Large-Scale Dense 3D Reconstruction with
  Loop Closure}. \bibinfo{publisher}{Springer International Publishing}.
\newblock pp. \bibinfo{pages}{500--516}.
\bibitem[{Le et~al.(2019)Le, Do, Hoang and Cheung}]{Sdrsac2019}
\bibinfo{author}{Le, H.M.}, \bibinfo{author}{Do, T.T.}, \bibinfo{author}{Hoang,
  T.}, \bibinfo{author}{Cheung, N.M.}, \bibinfo{year}{2019}.
\newblock \bibinfo{title}{Sdrsac: Semidefinite-based randomized approach for
  robust point cloud registration without correspondences}, in:
  \bibinfo{booktitle}{Proceedings of the IEEE/CVF Conference on Computer Vision
  and Pattern Recognition}, pp. \bibinfo{pages}{124--133}.
\bibitem[{Leordeanu and Hebert(2005)}]{PairwiseConstraints2005}
\bibinfo{author}{Leordeanu, M.}, \bibinfo{author}{Hebert, M.},
  \bibinfo{year}{2005}.
\newblock \bibinfo{title}{A spectral technique for correspondence problems
  using pairwise constraints}, in: \bibinfo{booktitle}{Tenth IEEE International
  Conference on Computer Vision (ICCV'05) Volume 1}, \bibinfo{publisher}{IEEE}.
\bibitem[{Li et~al.(2019)Li, Wu, Wang, Zhang, Xing and Yan}]{Siamrpn++2019}
\bibinfo{author}{Li, B.}, \bibinfo{author}{Wu, W.}, \bibinfo{author}{Wang, Q.},
  \bibinfo{author}{Zhang, F.}, \bibinfo{author}{Xing, J.},
  \bibinfo{author}{Yan, J.}, \bibinfo{year}{2019}.
\newblock \bibinfo{title}{Siamrpn++: Evolution of siamese visual tracking with
  very deep networks}, in: \bibinfo{booktitle}{Proceedings of the IEEE/CVF
  Conference on Computer Vision and Pattern Recognition}, pp.
  \bibinfo{pages}{4282--4291}.
\bibitem[{Liang et~al.(2021)Liang, Ji, Cheng, Chai, Wang and
  Ling}]{MultiLevelFeatureAggregation2021}
\bibinfo{author}{Liang, P.}, \bibinfo{author}{Ji, H.}, \bibinfo{author}{Cheng,
  E.}, \bibinfo{author}{Chai, Y.}, \bibinfo{author}{Wang, L.},
  \bibinfo{author}{Ling, H.}, \bibinfo{year}{2021}.
\newblock \bibinfo{title}{Learning local descriptors with multi-level feature
  aggregation and spatial context pyramid}.
\newblock \bibinfo{journal}{Neurocomputing} .
\bibitem[{Lin et~al.(2017)Lin, Dollár, Girshick, He, Hariharan and
  Belongie}]{FeaturePyramidNetworksforObjectDetection2017}
\bibinfo{author}{Lin, T.Y.}, \bibinfo{author}{Dollár, P.},
  \bibinfo{author}{Girshick, R.}, \bibinfo{author}{He, K.},
  \bibinfo{author}{Hariharan, B.}, \bibinfo{author}{Belongie, S.},
  \bibinfo{year}{2017}.
\newblock \bibinfo{title}{Feature pyramid networks for object detection}, in:
  \bibinfo{booktitle}{Proceedings of the IEEE conference on computer vision and
  pattern recognition}, pp. \bibinfo{pages}{2117--2125}.
\bibitem[{Lu et~al.(2019)Lu, Wan, Zhou, Fu, Yuan and Song}]{DeepVCP2019}
\bibinfo{author}{Lu, W.}, \bibinfo{author}{Wan, G.}, \bibinfo{author}{Zhou,
  Y.}, \bibinfo{author}{Fu, X.}, \bibinfo{author}{Yuan, P.},
  \bibinfo{author}{Song, S.}, \bibinfo{year}{2019}.
\newblock \bibinfo{title}{Deepvcp: An end-to-end deep neural network for point
  cloud registration}, in: \bibinfo{booktitle}{2019 IEEE/CVF International
  Conference on Computer Vision (ICCV)}, \bibinfo{publisher}{IEEE Computer
  Society}, \bibinfo{address}{Los Alamitos,CA}. pp. \bibinfo{pages}{12--21}.
\bibitem[{Lucas and Kanade(1981)}]{LucasKanade1981}
\bibinfo{author}{Lucas, B.D.}, \bibinfo{author}{Kanade, T.},
  \bibinfo{year}{1981}.
\newblock \bibinfo{title}{An iterative image registration technique with an
  application to stereo vision}, \bibinfo{publisher}{Vancouver, British
  Columbia}.
\bibitem[{Maron et~al.(2016)Maron, Dym, Kezurer, Kovalsky and
  Lipman}]{ConvexRelaxation2016}
\bibinfo{author}{Maron, H.}, \bibinfo{author}{Dym, N.},
  \bibinfo{author}{Kezurer, I.}, \bibinfo{author}{Kovalsky, S.},
  \bibinfo{author}{Lipman, Y.}, \bibinfo{year}{2016}.
\newblock \bibinfo{title}{Point registration via efficient convex relaxation}.
\newblock \bibinfo{journal}{ACM Transactions on Graphics (TOG)}
  \bibinfo{volume}{35}, \bibinfo{pages}{1--12}.
\bibitem[{Maturana and Scherer(2015)}]{VoxNet2015}
\bibinfo{author}{Maturana, D.}, \bibinfo{author}{Scherer, S.},
  \bibinfo{year}{2015}.
\newblock \bibinfo{title}{Voxnet: A 3d convolutional neural network for
  real-time object recognition}, in: \bibinfo{booktitle}{2015 IEEE/RSJ
  International Conference on Intelligent Robots and Systems (IROS)},
  \bibinfo{publisher}{IEEE}.
\bibitem[{Nair and Hinton(2010)}]{ReLU2010}
\bibinfo{author}{Nair, V.}, \bibinfo{author}{Hinton, G.E.},
  \bibinfo{year}{2010}.
\newblock \bibinfo{title}{Rectified linear units improve restricted boltzmann
  machines}, in: \bibinfo{booktitle}{Icml}.
\bibitem[{Pais et~al.(2020)Pais, Ramalingam, Govindu, Nascimento, Chellappa and
  Miraldo}]{3DRegNet2020}
\bibinfo{author}{Pais, G.D.}, \bibinfo{author}{Ramalingam, S.},
  \bibinfo{author}{Govindu, V.M.}, \bibinfo{author}{Nascimento, J.C.},
  \bibinfo{author}{Chellappa, R.}, \bibinfo{author}{Miraldo, P.},
  \bibinfo{year}{2020}.
\newblock \bibinfo{title}{3dregnet: A deep neural network for 3d point
  registration}, in: \bibinfo{booktitle}{2020 IEEE/CVF Conference on Computer
  Vision and Pattern Recognition (CVPR)}, \bibinfo{publisher}{IEEE},
  \bibinfo{address}{Piscataway,NJ}. pp. \bibinfo{pages}{7191--7201}.
\bibitem[{Perera and Barnes(2012)}]{MaximalCliques2012}
\bibinfo{author}{Perera, S.}, \bibinfo{author}{Barnes, N.},
  \bibinfo{year}{2012}.
\newblock \bibinfo{title}{Maximal cliques based rigid body motion segmentation
  with a rgb-d camera}, in: \bibinfo{editor}{Lee, K.M.},
  \bibinfo{editor}{Matsushita, Y.}, \bibinfo{editor}{Rehg, J.M.},
  \bibinfo{editor}{Hu, Z.} (Eds.), \bibinfo{booktitle}{Computer Vision – ACCV
  2012}, \bibinfo{publisher}{Springer Berlin Heidelberg}. pp.
  \bibinfo{pages}{120--133}.
\bibitem[{Pomerleau et~al.(2015)Pomerleau, Colas and
  Siegwart}]{ReviewPointCloudReg2015}
\bibinfo{author}{Pomerleau, F.}, \bibinfo{author}{Colas, F.},
  \bibinfo{author}{Siegwart, R.}, \bibinfo{year}{2015}.
\newblock \bibinfo{title}{A review of point cloud registration algorithms for
  mobile robotics}.
\newblock \bibinfo{journal}{Foundations and Trends in Robotics}
  \bibinfo{volume}{4}, \bibinfo{pages}{1--104}.
\bibitem[{Qi et~al.(2017)Qi, Yi, Su and Guibas}]{PointNetPlusPlus2017}
\bibinfo{author}{Qi, C.R.}, \bibinfo{author}{Yi, L.}, \bibinfo{author}{Su, H.},
  \bibinfo{author}{Guibas, L.J.}, \bibinfo{year}{2017}.
\newblock \bibinfo{title}{Pointnet++: deep hierarchical feature learning on
  point sets in a metric space}.
\bibitem[{Qiao et~al.(2020)Qiao, Liu, Suo, Wei, Shen and
  Wang}]{VitualCorresponmdences2020}
\bibinfo{author}{Qiao, Z.}, \bibinfo{author}{Liu, Z.}, \bibinfo{author}{Suo,
  C.}, \bibinfo{author}{Wei, H.}, \bibinfo{author}{Shen, Z.},
  \bibinfo{author}{Wang, H.}, \bibinfo{year}{2020}.
\newblock \bibinfo{title}{End-to-end 3d point cloud learning for registration
  task using virtual correspondences}, in: \bibinfo{booktitle}{2020 IEEE/RSJ
  International Conference on Intelligent Robots and Systems (IROS)},
  \bibinfo{publisher}{IEEE}, \bibinfo{address}{Piscataway,NJ}. pp.
  \bibinfo{pages}{2678--2683}.
\bibitem[{Riegler et~al.(2017)Riegler, Osman~Ulusoy and Geiger}]{Octnet2017}
\bibinfo{author}{Riegler, G.}, \bibinfo{author}{Osman~Ulusoy, A.},
  \bibinfo{author}{Geiger, A.}, \bibinfo{year}{2017}.
\newblock \bibinfo{title}{Octnet: Learning deep 3d representations at high
  resolutions}, in: \bibinfo{booktitle}{Proceedings of the IEEE conference on
  computer vision and pattern recognition}, pp. \bibinfo{pages}{3577--3586}.
\bibitem[{Rosen et~al.(2019)Rosen, Carlone, Bandeira and Leonard}]{SE-Sync2019}
\bibinfo{author}{Rosen, D.M.}, \bibinfo{author}{Carlone, L.},
  \bibinfo{author}{Bandeira, A.S.}, \bibinfo{author}{Leonard, J.J.},
  \bibinfo{year}{2019}.
\newblock \bibinfo{title}{Se-sync: A certifiably correct algorithm for
  synchronization over the special euclidean group}.
\newblock \bibinfo{journal}{The International Journal of Robotics Research}
  \bibinfo{volume}{38}, \bibinfo{pages}{95--125}.
\newblock \URLprefix
  \url{https://dspace.mit.edu/bitstream/1721.1/120180/1/1612.07386.pdf}.
\bibitem[{Rusinkiewicz and Levoy(2001)}]{EfficientVariantsofICP2001}
\bibinfo{author}{Rusinkiewicz, S.}, \bibinfo{author}{Levoy, M.},
  \bibinfo{year}{2001}.
\newblock \bibinfo{title}{Efficient variants of the icp algorithm}, in:
  \bibinfo{booktitle}{Proceedings Third International Conference on 3-D Digital
  Imaging and Modeling}, \bibinfo{publisher}{IEEE Comput. Soc}.
\bibitem[{Schops et~al.(2019)Schops, Sattler and Pollefeys}]{BadSlam2019}
\bibinfo{author}{Schops, T.}, \bibinfo{author}{Sattler, T.},
  \bibinfo{author}{Pollefeys, M.}, \bibinfo{year}{2019}.
\newblock \bibinfo{title}{Bad slam: Bundle adjusted direct rgb-d slam}, in:
  \bibinfo{booktitle}{Proceedings of the IEEE/CVF Conference on Computer Vision
  and Pattern Recognition}, pp. \bibinfo{pages}{134--144}.
\bibitem[{Su et~al.(2015)Su, Maji, Kalogerakis and Learned-Miller}]{MVCNN2015}
\bibinfo{author}{Su, H.}, \bibinfo{author}{Maji, S.},
  \bibinfo{author}{Kalogerakis, E.}, \bibinfo{author}{Learned-Miller, E.},
  \bibinfo{year}{2015}.
\newblock \bibinfo{title}{Multi-view convolutional neural networks for 3d shape
  recognition}, in: \bibinfo{booktitle}{Proceedings of the IEEE international
  conference on computer vision}, pp. \bibinfo{pages}{945--953}.
\bibitem[{Thomas et~al.(2019)Thomas, Qi, Deschaud, Marcotegui, Goulette and
  Guibas}]{Kpconv2019}
\bibinfo{author}{Thomas, H.}, \bibinfo{author}{Qi, C.R.},
  \bibinfo{author}{Deschaud, J.E.}, \bibinfo{author}{Marcotegui, B.},
  \bibinfo{author}{Goulette, F.}, \bibinfo{author}{Guibas, L.J.},
  \bibinfo{year}{2019}.
\newblock \bibinfo{title}{Kpconv: Flexible and deformable convolution for point
  clouds}, in: \bibinfo{booktitle}{Proceedings of the IEEE/CVF International
  Conference on Computer Vision}, pp. \bibinfo{pages}{6411--6420}.
\bibitem[{Wang et~al.(2017)Wang, Liu, Guo, Sun and Tong}]{O-CNN2017}
\bibinfo{author}{Wang, P.S.}, \bibinfo{author}{Liu, Y.}, \bibinfo{author}{Guo,
  Y.X.}, \bibinfo{author}{Sun, C.Y.}, \bibinfo{author}{Tong, X.},
  \bibinfo{year}{2017}.
\newblock \bibinfo{title}{O-cnn}.
\newblock \bibinfo{journal}{ACM Transactions on Graphics} \bibinfo{volume}{36},
  \bibinfo{pages}{1--11}.
\newblock \URLprefix \url{http://arxiv.org/pdf/1712.01537}.
\bibitem[{Wang and Solomon(2019)}]{DeepClosetPoint2019}
\bibinfo{author}{Wang, Y.}, \bibinfo{author}{Solomon, J.M.},
  \bibinfo{year}{2019}.
\newblock \bibinfo{title}{Deep closest point: Learning representations for
  point cloud registration}, in: \bibinfo{booktitle}{2019 IEEE/CVF
  International Conference on Computer Vision (ICCV)}, \bibinfo{publisher}{IEEE
  Computer Society}, \bibinfo{address}{Los Alamitos,CA}. pp.
  \bibinfo{pages}{3522--3531}.
\bibitem[{Wang et~al.(2019)Wang, Sun, Liu, Sarma, Bronstein and
  Solomon}]{DGCNN2019}
\bibinfo{author}{Wang, Y.}, \bibinfo{author}{Sun, Y.}, \bibinfo{author}{Liu,
  Z.}, \bibinfo{author}{Sarma, S.E.}, \bibinfo{author}{Bronstein, M.M.},
  \bibinfo{author}{Solomon, J.M.}, \bibinfo{year}{2019}.
\newblock \bibinfo{title}{Dynamic graph cnn for learning on point clouds}.
\newblock \bibinfo{journal}{ACM Transactions on Graphics} \bibinfo{volume}{38},
  \bibinfo{pages}{1--12}.
\bibitem[{Wei et~al.(2020)Wei, Yu and Sun}]{View-GCN2020}
\bibinfo{author}{Wei, X.}, \bibinfo{author}{Yu, R.}, \bibinfo{author}{Sun, J.},
  \bibinfo{year}{2020}.
\newblock \bibinfo{title}{View-gcn: View-based graph convolutional network for
  3d shape analysis}, in: \bibinfo{booktitle}{Proceedings of the IEEE/CVF
  Conference on Computer Vision and Pattern Recognition}, pp.
  \bibinfo{pages}{1850--1859}.
\bibitem[{Whelan et~al.(2015)Whelan, Leutenegger, Salas-Moreno, Glocker and
  Davison}]{ElasticFusion2015}
\bibinfo{author}{Whelan, T.}, \bibinfo{author}{Leutenegger, S.},
  \bibinfo{author}{Salas-Moreno, R.}, \bibinfo{author}{Glocker, B.},
  \bibinfo{author}{Davison, A.}, \bibinfo{year}{2015}.
\newblock \bibinfo{title}{Elasticfusion: Dense slam without a pose graph},
  \bibinfo{publisher}{Robotics: Science and Systems}.
\bibitem[{Wold et~al.(1987)Wold, Esbensen and Geladi}]{PCA1987}
\bibinfo{author}{Wold, S.}, \bibinfo{author}{Esbensen, K.},
  \bibinfo{author}{Geladi, P.}, \bibinfo{year}{1987}.
\newblock \bibinfo{title}{Principal component analysis}.
\newblock \bibinfo{journal}{Chemometrics and Intelligent Laboratory Systems}
  \bibinfo{volume}{2}, \bibinfo{pages}{37--52}.
\newblock \URLprefix
  \url{https://www.sciencedirect.com/science/article/pii/0169743987800849}.
\bibitem[{Wu et~al.(2015)Wu, Song, Khosla, Yu, Zhang, Tang and
  Xiao}]{3Dshapenets2015}
\bibinfo{author}{Wu, Z.}, \bibinfo{author}{Song, S.}, \bibinfo{author}{Khosla,
  A.}, \bibinfo{author}{Yu, F.}, \bibinfo{author}{Zhang, L.},
  \bibinfo{author}{Tang, X.}, \bibinfo{author}{Xiao, J.}, \bibinfo{year}{2015}.
\newblock \bibinfo{title}{3d shapenets: A deep representation for volumetric
  shapes}, in: \bibinfo{booktitle}{Proceedings of the IEEE conference on
  computer vision and pattern recognition}, pp. \bibinfo{pages}{1912--1920}.
\bibitem[{Yang et~al.(2021)Yang, Shi and Carlone}]{TEASER2021}
\bibinfo{author}{Yang, H.}, \bibinfo{author}{Shi, J.},
  \bibinfo{author}{Carlone, L.}, \bibinfo{year}{2021}.
\newblock \bibinfo{title}{Teaser: Fast and certifiable point cloud
  registration}.
\newblock \bibinfo{journal}{IEEE Transactions on Robotics}
  \bibinfo{volume}{37}, \bibinfo{pages}{314--333}.
\bibitem[{Yang et~al.(2016)Yang, Li, Campbell and Jia}]{GoICP2016}
\bibinfo{author}{Yang, J.}, \bibinfo{author}{Li, H.},
  \bibinfo{author}{Campbell, D.}, \bibinfo{author}{Jia, Y.},
  \bibinfo{year}{2016}.
\newblock \bibinfo{title}{Go-icp: A globally optimal solution to 3d icp
  point-set registration}.
\newblock \bibinfo{journal}{IEEE Transactions on Pattern Analysis and Machine
  Intelligence} \bibinfo{volume}{38}, \bibinfo{pages}{2241--2254}.
\bibitem[{Yu et~al.(2018)Yu, Meng and Yuan}]{MHBN2018}
\bibinfo{author}{Yu, T.}, \bibinfo{author}{Meng, J.}, \bibinfo{author}{Yuan,
  J.}, \bibinfo{year}{2018}.
\newblock \bibinfo{title}{Multi-view harmonized bilinear network for 3d object
  recognition}, in: \bibinfo{booktitle}{Proceedings of the IEEE Conference on
  Computer Vision and Pattern Recognition}, pp. \bibinfo{pages}{186--194}.
\bibitem[{Zeng et~al.(2017)Zeng, Song, Niessner, Fisher, Xiao and
  Funkhouser}]{3DMatch2017}
\bibinfo{author}{Zeng, A.}, \bibinfo{author}{Song, S.},
  \bibinfo{author}{Niessner, M.}, \bibinfo{author}{Fisher, M.},
  \bibinfo{author}{Xiao, J.}, \bibinfo{author}{Funkhouser, T.},
  \bibinfo{year}{2017}.
\newblock \bibinfo{title}{3dmatch: Learning local geometric descriptors from
  rgb-d reconstructions}.
\bibitem[{Zhou et~al.(2016)Zhou, Park and Koltun}]{FastGlobalRegistration2016}
\bibinfo{author}{Zhou, Q.Y.}, \bibinfo{author}{Park, J.},
  \bibinfo{author}{Koltun, V.}, \bibinfo{year}{2016}.
\newblock \bibinfo{title}{Fast global registration}, in:
  \bibinfo{booktitle}{Computer Vision – ECCV 2016},
  \bibinfo{publisher}{Springer International Publishing}. pp.
  \bibinfo{pages}{766--782}.
\bibitem[{Zhou et~al.(2018)Zhou, Park and Koltun}]{Open3D2018}
\bibinfo{author}{Zhou, Q.Y.}, \bibinfo{author}{Park, J.},
  \bibinfo{author}{Koltun, V.}, \bibinfo{year}{2018}.
\newblock \bibinfo{title}{Open3d: A modern library for 3d data processing}.
\newblock \URLprefix \url{https://arxiv.org/pdf/1801.09847.pdf}.

\end{thebibliography}
